\documentclass{article}

     \PassOptionsToPackage{numbers, compress}{natbib}

\usepackage[preprint]{neurips_2026}

\usepackage[utf8]{inputenc} 
\usepackage[T1]{fontenc}    
\usepackage{hyperref}       
\usepackage{url}            
\usepackage{booktabs}       
\usepackage{amsfonts}       
\usepackage{nicefrac}       
\usepackage{microtype}      
\usepackage{xcolor}         

\usepackage{microtype}
\usepackage{graphicx}
\usepackage{subfigure}
\usepackage{booktabs} 
\usepackage{longtable}
\usepackage{multirow}
\usepackage{makecell}
\usepackage[ruled]{algorithm2e}
\usepackage{tabularx}
\usepackage{array}
\usepackage{placeins}
\usepackage{wrapfig}

\usepackage{amsmath}
\usepackage{amssymb}
\usepackage{mathtools}
\usepackage{amsthm}

\theoremstyle{plain}

\theoremstyle{definition}

\theoremstyle{remark}

\usepackage[textsize=tiny]{todonotes}

\title{FiLoRA: Focus-and-Ignore LoRA for \\Controllable Feature Reliance}

\newcounter{eqfn}
\newcounter{secfn}
\newcounter{corrfn}

\author{%
  Hyunsuk Chung$^{1}$\thanks{Equal contribution.}\setcounter{eqfn}{\value{footnote}} \And
  Soyeon Caren Han$^{1}$\footnotemark[\value{eqfn}]\hspace{5pt}\thanks{Corresponding authors: Soyeon Caren Han \texttt{<caren.han@unimelb.edu.au>}, Kyungreem Han \texttt{<khan@kist.re.kr>}}\setcounter{corrfn}{\value{footnote}} \And
  Seungyeon Ji$^{2,3}$\thanks{These authors contributed equally as second authors.}\setcounter{secfn}{\value{footnote}} \And
  Jinwoo Kim$^{1}$\footnotemark[\value{secfn}] \AND
  Eun-Jung Holden$^{1}$ \quad
  Kyungreem Han$^{2,4}$\footnotemark[\value{corrfn}] \\
  \mdseries
  $^1$The University of Melbourne, Melbourne, Australia \\
  $^2$Korea Institute of Science and Technology, Seoul, Republic of Korea \\
  $^3$Korea University, Seoul, Republic of Korea\\
  $^4$University of Science and Technology KIST School, Seoul, Republic of Korea \\
  \texttt{\{hyunsuk.chung.1, caren.han, jinwoo.kim.1, eunjung.holden\}@unimelb.edu.au} \\
  \texttt{\{syji, khan\}@kist.re.kr}
}

\begin{document}

\maketitle

\begin{abstract}
Multimodal foundation models integrate heterogeneous signals across modalities, yet it remains unclear whether their predictions can be controlled by explicitly modulating reliance on different internal feature pathways. Existing approaches to shortcut and spurious behavior primarily rely on post hoc analysis or data-level interventions, offering limited ability to directly intervene on how models use information.
We introduce \textbf{FiLoRA (Focus-and-Ignore LoRA)}, an instruction-conditioned, parameter-efficient adaptation framework that enables controllable modulation of feature reliance while preserving a common prediction task and label space. FiLoRA organizes adaptation into feature-aligned low-rank modules and applies instruction-conditioned gating, allowing natural language instructions to regulate the contributions of trainable adaptation pathways.
We evaluate FiLoRA across both controlled classification settings and generative multimodal tasks, and under a range of instruction types, including natural and compositional instructions. Results show that FiLoRA induces consistent and interpretable shifts in feature reliance, selectively amplifying or suppressing different adaptation groups in accordance with the instruction, without altering task semantics.
Our findings suggest that instruction-conditioned parameter adaptation can serve as a practical mechanism for modulating feature reliance, providing a new perspective on controllability and analysis of multimodal systems beyond output-level prompting or post hoc interpretation.
\end{abstract}

\begin{figure}[t]
    \centering
    \includegraphics[width=\columnwidth]{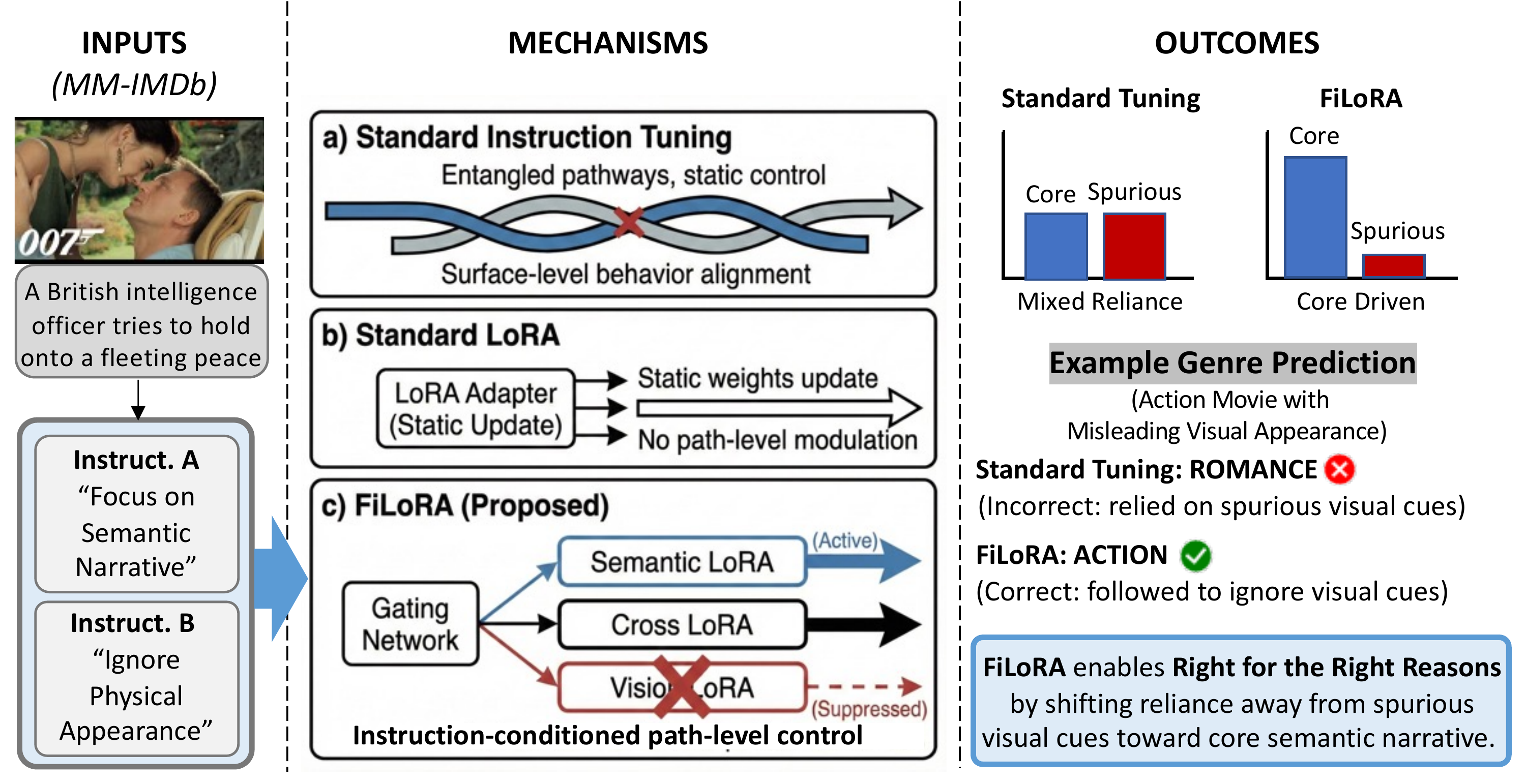}
    \vspace{-1em}
    \caption{Feature reliance control in multimodal models.
(a) Standard instruction tuning aligns surface-level outputs without explicitly modulating internal computation.
(b) Standard LoRA applies uniform parameter updates without pathway-level control.
(c) \textbf{FiLoRA (ours)} enables instruction-conditioned modulation by gating feature-aligned LoRA modules.
Given different natural language instructions (e.g. focusing on semantic content or ignoring visual appearance), FiLoRA selectively amplifies or suppresses corresponding feature pathways, enabling the same model to perform the same task under different reliance conditions.}
\vspace{-1.5em}
    \label{fig:FiLoRA}
\end{figure}

\section{Introduction}

Understanding multimodal models requires not only observing their outputs, but also controlling how they use different sources of information. As noted by Feynman, \emph{``What I cannot create, I do not understand.''} In this spirit, we investigate whether internal feature reliance in multimodal models can be explicitly controlled, rather than merely analyzed post hoc. While modern multimodal foundation models can process heterogeneous inputs such as text, vision, and audio~\cite{qwen, gemma3n, omnivinci, phi}, it remains unclear whether their predictions can be systematically altered by intervening on specific internal feature pathways, rather than changing the task itself.
In many multimodal tasks, correct predictions can be obtained by exploiting superficial or shortcut cues that correlate with labels in the training distribution, rather than relying on task-relevant semantic information~\cite{whatmakes, looklisten, ijcai2025p280}. For example, movie genre classifiers may rely on visual cues such as poster color or typography that correlate with labels but are not semantically central to the task. While such behavior can be partially revealed through post hoc analysis techniques~\cite{geometric, lidevil}, these approaches only provide retrospective explanations and do not enable explicit intervention on feature usage.

A natural alternative is to use instructions to control model behavior. However, existing approaches operate primarily at the level of output behavior. Instruction tuning and prompting can alter surface-level predictions, but do not guarantee corresponding changes in internal computation~\cite{fromlmtoit, instructionedit, inststeering}. Recent methods such as FIT~\cite{fit} evaluate whether models follow feature-level guidance expressed in instructions, but assess control through output behavior and do not provide mechanisms to directly modulate internal computation paths. Similarly, FocalLoRA~\cite{focallora} allocates adaptation capacity based on instruction hierarchy, but focuses on behavioral alignment rather than selectively activating or suppressing feature-specific computation paths under a fixed task.
This exposes a fundamental gap: existing methods either analyze feature reliance retrospectively, or influence outputs without controlling the underlying computation. As a result, it remains unclear whether feature usage itself can be explicitly and reliably controlled.

\vspace{-0.2em}

This leaves a central question unanswered: \textit{can natural language instructions be used to explicitly control internal feature reliance in multimodal models, without changing the task definition, label space, or base representations?}
We propose \textbf{FiLoRA (Focus-and-Ignore LoRA)}, an instruction-conditioned, parameter-efficient adaptation framework that enables explicit control over internal feature reliance in multimodal models. FiLoRA extends LoRA by decomposing adaptation parameters into groups corresponding to distinct functional computation paths, such as semantic reasoning, visual appearance processing, and acoustic feature extraction. Natural language instructions are encoded and mapped to gate values that modulate the contribution of each LoRA group during the forward pass. By keeping the base model frozen and applying instruction-conditioned gating only to LoRA updates, FiLoRA enables soft, sample-wise intervention on internal computation paths while preserving the original task and representations.
We evaluate FiLoRA across both controlled classification settings and generative multimodal tasks, and under a range of instruction types, including natural and compositional instructions (e.g., \emph{``describe the main action without referring to visual style''}, \emph{``focus on interactions between people while ignoring background cues''}). Results show that FiLoRA induces consistent and interpretable shifts in feature usage, selectively amplifying or suppressing different feature groups in accordance with the instruction, while preserving task semantics.
Our contributions can be summarized in three parts:
\begin{enumerate}
    \item We formalize instruction following as a problem of \emph{controlled intervention on feature reliance}, distinguishing it from output-level behavior control.
    \item We introduce FiLoRA, a grouped and gated LoRA framework that enables differentiable, instruction-conditioned modulation of internal computation paths.
    \item We demonstrate across multimodal benchmarks, including generative settings, that FiLoRA achieves consistent and interpretable control over feature usage under diverse instructions.
\end{enumerate}

\section{Related Work}

\paragraph{Multimodal Instruction Following.}\hspace{-8pt}
Recent multimodal foundation models \cite{llava_next, instructblip, qwen2_vl, kosmos2, guide} have further explored instruction-guided multimodal reasoning and controllable vision--language behavior, yet such approaches remain focused on output alignment rather than explicit regulation of internal feature reliance. Related efforts on multimodal attention control or prompting~\cite{attention_control, visual_prompting} primarily modulate attention weights or input saliency, but do not provide explicit mechanisms to regulate feature-level reliance within internal computation.
Multimodal models may exploit superficial correlations that are predictive under the training distribution but semantically incidental to the task~\cite{whatmakes, looklisten, ijcai2025p280}.
This phenomenon aligns with prior analyses of shortcut learning and spurious correlations, which show that models often exploit predictive but semantically irrelevant cues when such shortcuts are available~\cite{geirhos_shortcut_learning, shortcut_llm, spurious_benchmm}.
Such spurious reliance, including poster color in movie genre classification, background patterns in emotion recognition, or speaker identity cues in audio--visual datasets, has been repeatedly observed in benchmarks; MM-IMDb \cite{arevalo2017gatedmultimodalunitsinformation}, CREMA-D \cite{cremad}, and RAVDESS \cite{ravdess}.
\vspace{-1.0em}
\paragraph{Feature Steering and Bias Mitigation.}
In the text-only domain, recent work has explored feature steering, representation editing, and controllable generation \cite{fit, controllingcav, repres_edit, tosteernotsteer, turner2023steering, zou2023representation}, demonstrating that certain behaviors can be influenced via latent or activation-level control. FIT \cite{fit} steers language models via feature-level instructions, but is limited to textual features and does not generalize to multimodal reliance control.
These methods neither distinguish nor selectively suppress visual or acoustic cues, nor do they provide mechanisms to separate core semantic information from spurious modality-specific correlations.
Existing multimodal bias mitigation methods~\cite{assume, multimodal-representation-learning, geometric} operate primarily at training time and aim to reduce dataset-level correlations, whereas our setting requires test-time, instruction-conditioned control over feature reliance without retraining.
Consequently, they cannot accommodate user-specified directives such as ``ignore the background'' or ``do not use speaker identity'' during inference.
This lack of controllability limits their applicability in interactive or safety-critical scenarios.
Parameter-efficient fine-tuning methods such as LoRA \cite{lora}, QLoRA \cite{qlora}, adapters \cite{adapter}, and FocalLoRA \cite{focallora} enable efficient adaptation of large models but rely on static or pre-defined update pathways. While FocalLoRA allocates adaptation capacity unevenly across model components, these methods do not support instruction-conditioned routing or selective activation of feature-specific computation paths, particularly in multimodal settings.
Knowledge editing methods such as ROME and MEMAT enable targeted parameter modifications at test time, but focus on altering factual content or stored knowledge rather than regulating feature reliance under fixed task semantics~\cite{meng_knowledge_edit, attention_edit}.
\vspace{-0.8em}
\paragraph{Gap and Motivation.}
Prior work lacks a unified mechanism to translate natural language specifications of feature roles into explicit, instruction-conditioned control over internal multimodal computation paths.
This gap motivates our work.
We argue that enabling instruction-conditioned selective multimodal reasoning requires a mechanism that translates natural language specifications of feature roles into explicit control over internal computation paths.
Rather than redefining the prediction task, our approach uses instruction-conditioned adaptation and training-time proxy supervision to modulate feature reliance.
This perspective directly leads to our proposed framework, which operationalizes feature-level instructions as instruction-conditioned routing over parameter-efficient adaptation pathways.

\section{FiLoRA}
\label{sec:method}

\subsection{Problem Formulation}
\label{sec:prob_form}
We consider a multimodal prediction task with a fixed label space $\mathcal{Y}$. 
Each input sample $x_i$ consists of heterogeneous modalities (e.g., text, vision, audio) and is associated with a target label $y_i \in \mathcal{Y}$. 
The task definition, label space, and evaluation objective remain invariant throughout. Our goal is not to change what the model predicts, but to control \emph{how} predictions are formed specifically, which internal feature pathways the model relies on.
To enable controlled analysis of feature reliance, we introduce an auxiliary training signal that emphasizes different sources of evidence (e.g., semantic or appearance-based features) while remaining within the same label space. 
This signal is used only during training as an intervention mechanism and is never used as an evaluation target.
Further details of its construction are provided in Appendix~\ref{appendix:proxy_labels}.
This setup allows us to isolate changes in feature reliance without altering task semantics or the label space.

\subsection{Instruction as Feature-Level Control}

Each input is paired with a natural language instruction $I_i$. 
Rather than redefining the task, instructions specify how different feature types should be used when forming a prediction (e.g., emphasizing semantic content or suppressing visual appearance).
Instructions serve as the sole control signal provided to the model. 
Formally, each instruction is encoded into a continuous representation:
\[
\mathbf{z}_i = h_\phi(I_i),
\]
where $h_\phi(\cdot)$ denotes an instruction encoder.
This representation parameterizes the instruction-conditioned gating function (Section~\ref{sec:gating}), determining how strongly different feature pathways are emphasized or suppressed during prediction.
This design allows instruction-following to be interpreted as a controlled intervention: the task and label space remain fixed, while internal feature usage varies according to instruction semantics.
Importantly, instructions modulate internal computation rather than directly constraining outputs.

\subsection{Base Model and Parameter-Efficient Adaptation}

Let $f_{\theta}$ denote a pretrained multimodal model with parameters $\theta$. 
We freeze all base parameters to avoid representation drift and confounding effects introduced by full fine-tuning.
Adaptation is performed using low-rank updates in the spirit of LoRA, which provide a minimal and structured way to intervene on model parameters. 
For a linear transformation with weight matrix $W \in \mathbb{R}^{d_{\text{out}} \times d_{\text{in}}}$, LoRA introduces an additive update:
\[
\Delta W = B A^\top,
\]
where $A \in \mathbb{R}^{d_{\text{in}} \times r}$ and $B \in \mathbb{R}^{d_{\text{out}} \times r}$ with $r \ll \min(d_{\text{in}}, d_{\text{out}})$. 
The adapted transformation is given by $W' = W + \Delta W$.
While effective for parameter-efficient adaptation, standard LoRA applies a single undifferentiated update per layer. 
As a result, feature contributions remain entangled within the same update, making it difficult to selectively control reliance on different computation paths. 
This motivates the grouped formulation introduced in the next section.
We apply LoRA updates to all linear projections in attention and feed-forward blocks unless otherwise specified.


\subsection{Grouped LoRA for Path-Level Control}
\label{sec:grouped_lora}

To enable controllable modulation of feature reliance, we organize trainable LoRA updates into groups associated with task-relevant evidence categories. In our primary setting, these categories are defined using interpretable feature taxonomies, including semantic content, visual appearance, acoustic properties, and background information. The resulting groups provide distinct adaptation pathways through which the model can learn to emphasize or suppress different sources of evidence, without modifying the frozen backbone.
Each adapted transformation is written as:
\[
W' = W + \sum_{g \in \mathcal{G}} \Delta W_g,
\]
where each group-specific update $\Delta W_g$ is parameterized as a low-rank adaptation, $\Delta W_g = B_g A_g^\top$, with rank $r \ll \min(d_{\text{in}}, d_{\text{out}})$.
Unlike standard LoRA, which applies a single undifferentiated update per adapted transformation, FiLoRA maintains multiple group-specific updates that can be modulated independently. The group definitions remain fixed during training, while their contributions are controlled by instruction-conditioned gates.
Importantly, the groups represent evidence categories rather than instruction templates. A single natural language instruction can therefore modulate multiple adaptation groups simultaneously, supporting compositional control over different sources of evidence. In this work, we primarily employ semantically aligned groups to facilitate interpretable evaluation, and examine the sensitivity of FiLoRA to alternative grouping strategies in Section~5.5.

\subsection{Instruction-Conditioned Gating}
\label{sec:gating}

While grouped LoRA provides distinct trainable adaptation pathways, grouping alone does not determine how strongly each pathway should contribute under a particular instruction. To enable instruction-dependent modulation, we introduce an instruction-conditioned gating mechanism. Each instruction $I_i$ is encoded into a continuous representation and mapped to a gate vector:
\[
\mathbf{g}(I_i) = \sigma\!\left(h_\phi(I_i)\right)
\in [0,1]^{|\mathcal{G}|},
\]
where $h_\phi(\cdot)$ denotes an instruction encoder followed by a linear projection, and $\sigma(\cdot)$ is an element-wise sigmoid. The resulting gates are shared across layers and applied consistently to the corresponding group-specific LoRA updates.
For a given instruction, the adapted transformation is:
\[
W' = W + \sum_{g \in \mathcal{G}}
g_g(I_i)\,\Delta W_g,
\]
where each gate $g_g(I_i)$ continuously scales the contribution of the corresponding adaptation group $\Delta W_g$. This formulation enables instructions to modulate multiple adaptation pathways simultaneously, rather than selecting a single discrete module.
The gates depend exclusively on the instruction text and are not conditioned on supervision targets or intermediate model states. For autoregressive generation, the gate vector is computed once per instruction and remains fixed across decoding steps. Although the gate values remain unchanged during generation, the contributions of the gated LoRA updates to the evolving hidden representations may vary across decoding steps.
By modulating trainable LoRA updates while keeping the pretrained backbone frozen, FiLoRA provides explicit instruction-conditioned control over adaptation pathways. We evaluate whether this modulation produces corresponding changes in functional feature reliance through gate statistics, prediction sensitivity, and targeted feature interventions.



\subsection{Controlled Intervention Training}
To associate instructions with distinct reliance patterns, we introduce a controlled training signal that biases the model toward different feature types while keeping the task fixed.
Concretely, during training, the supervision signal is varied to emphasize different sources of evidence (e.g., semantic or appearance-based cues), while always remaining within the same label space. 
This signal is used only as a training-time intervention and is not provided to the model as input.
The model observes only the input modalities and the instruction text. 
Therefore, any learned shift in feature reliance must emerge from the interaction between instruction-conditioned gating and the optimization objective, rather than explicit conditioning on training signals.
At evaluation time, all predictions are assessed with respect to the original task labels, ensuring that observed effects reflect changes in feature reliance rather than task redefinition.

\subsection{Training Objective}

We optimize FiLoRA using a classification loss with condition-dependent supervision targets. For each training sample, let $y_i$ denote the original task label and $y_i^{(s)}$ denote a spurious proxy label constructed from shortcut-associated evidence (Appendix~\ref{appendix:proxy_labels}). The training condition $c_i$ determines the supervision target:
\[
y_i^{\mathrm{target}} =
\begin{cases}
y_i, & \text{if } c_i \text{ emphasizes core evidence},\\
y_i^{(s)}, & \text{if } c_i \text{ emphasizes spurious evidence}.
\end{cases}
\]
The classification loss is
\[
\mathcal{L}_{\mathrm{cls}}^{(i)}
= \mathrm{CE}\big(f(x_i,I_i),y_i^{\mathrm{target}}\big).
\]
Both targets belong to the original label space, and the same loss formulation is used across training conditions. The condition determines the supervision target but is not provided as an additional model input.
To encourage instruction-aligned modulation of adaptation groups, we optionally introduce a weak gate regularization term:
\[
\mathcal{L}_{\mathrm{gate}}^{(i)}
= \sum_{g \in \mathcal{G}} \alpha_g^{(i)}\,g_g(I_i),
\]
where $\alpha_g^{(i)} \in \{-1,0,+1\}$ specifies the intended direction of gate modulation: $-1$ encourages higher gate activation, $+1$ encourages suppression, and $0$ imposes no preference. The overall objective is
\[
\mathcal{L}^{(i)}
= \mathcal{L}_{\mathrm{cls}}^{(i)}
+ \lambda\mathcal{L}_{\mathrm{gate}}^{(i)},
\]
where $\lambda$ controls the strength of the regularization. This term provides a soft inductive bias toward instruction-aligned gating without imposing hard constraints on predictions. At evaluation time, predictions are assessed exclusively against the original task labels.

\subsection{Mechanism Interpretation}
\vspace{-0.5em}
Under this formulation, instructions modulate the contributions of trainable LoRA adaptation pathways without redefining the prediction task. The model learns to associate instruction semantics with distinct patterns of feature usage under a common loss formulation, although supervision targets may vary during training. These changes are assessed through gate statistics, prediction sensitivity, and targeted feature interventions, enabling systematic analysis of feature reliance under fixed task semantics.

\paragraph{Why Parameter-Level Control.}
\vspace{-1em}
Unlike output-level prompting or activation steering, FiLoRA directly modulates trainable low-rank parameter updates during the forward pass while keeping the pretrained backbone frozen. This provides structured control over adaptation pathways without modifying the backbone representations.

\paragraph{Why This Is Not Multi-Task Learning.}
\vspace{-1em}
FiLoRA maintains a common prediction task and fixed label space across all conditions. Although supervision targets may differ during training, the loss formulation and model architecture remain unchanged. Proxy supervision serves only as a training-time intervention, while evaluation consistently uses the original task labels.

\section{Evaluation Setup}
We evaluate whether FiLoRA enables controlled and interpretable modulation of feature reliance under fixed task semantics. 
While maintaining competitive task performance, our primary goal is to assess whether natural language instructions can reliably steer internal computation and induce consistent functional changes in model behavior.
To evaluate both controllability and generalization, we include a mixture of controlled and natural multimodal settings. 
Controlled settings allow precise attribution of feature-level effects with minimal confounding factors, while more natural settings evaluate whether instruction-conditioned reliance modulation extends to open-ended generation.
Accordingly, our evaluation includes both classification tasks and generative multimodal settings such as image captioning.
All experiments follow a unified training configuration, with implementation details provided in Appendix~\ref{app:implementation_details}. 
Instruction datasets contain both template-based and natural language instructions of varying complexity, including paraphrased and compositional forms. 
For each dataset, we generate approximately 1,000 instructions using frozen multimodal foundation models; additional details are provided in Appendix~\ref{app:instruction_dataset}.

\paragraph{Dataset Description}
\vspace{-0.5em}
We select datasets with identifiable feature groups and fixed task semantics to enable controlled analysis of instruction-conditioned feature reliance; detailed criteria are provided in Appendix~\ref{appendix:dataset_criteria}.
\textbf{1) Tiny MM-IMDb} is included as a controlled vision--language setting for precise feature attribution. 
Movie plots contain semantic and narrative information, while posters exhibit separable visual attributes such as color, typography, and layout. 
These properties enable selective emphasis or suppression of feature groups through instructions.
To verify that observed controllability is not an artifact of scale, we additionally evaluate FiLoRA on larger-capacity backbones.
\textbf{2) CREMA-D}~\cite{cremad} contains emotional expressions from a diverse set of actors with controlled variation across identity-related attributes, supporting identifiable audio--visual feature groups. 
This structure enables analysis of whether reliance modulation generalizes beyond specific identities while reducing identity-specific shortcuts.
\textbf{3) RAVDESS}~\cite{ravdess} provides balanced spoken and sung emotional speech with systematically controllable acoustic properties. 
Its balanced recording setup minimizes spurious correlations and supports analysis of acoustic feature reliance under controlled conditions.
To evaluate whether FiLoRA generalizes beyond classification settings, we additionally conduct generative evaluation on \textbf{4) Flickr30K}~\cite{flickr30k}, where instructions specify which semantic or visual attributes should be emphasized or suppressed during image caption generation.

\vspace{-1em}
\paragraph{Backbone Models}
We use dataset-appropriate multimodal backbones to avoid modality mismatches.
For vision--language experiments, we use Qwen2.5-VL (7B, 32B) and Gemma-3n-E4B. 
For audio--visual experiments, we use Qwen2.5-Omni-7B alongside AV-HuBERT~\cite{avhubert} and AudioCLIP~\cite{audioclip}.
This setup allows us to evaluate whether instruction-conditioned reliance modulation generalizes across model scales, modalities, and architectural inductive biases. 
Additional implementation details are provided in Appendix~\ref{sec:appen_tech_ref}.

\vspace{-1em}

\paragraph{Metrics}
We evaluate instruction-conditioned controllability using two complementary metrics.
\emph{1) Gate Modulation Range (GMR)} measures the magnitude of gate changes between contrasting feature-level instructions:
\[
\mathrm{GMR}_g =
\mathbb{E}_{(I^{\mathrm{focus}},I^{\mathrm{ignore}})}
\left[
\left|g_g(I^{\mathrm{focus}})-g_g(I^{\mathrm{ignore}})\right|
\right].
\]
Here, $I^{\mathrm{focus}}$ and $I^{\mathrm{ignore}}$ are paired instructions that emphasize and suppress feature group $g$, respectively.
\emph{2) Reliance Sensitivity (RS)} measures prediction sensitivity to perturbations of feature-group gates:
\[
\mathrm{RS}_g =
\mathbb{E}_{(x,I,y)}
\left[
\left|
\frac{\partial \log p(y \mid x,I)}
{\partial g_g(I)}
\right|
\right].
\]
RS averages the absolute gradient of the target log-probability with respect to each group-specific gate across evaluation samples. Together, GMR and RS characterize instruction-induced gate modulation and its functional influence on predictions.
In addition, we report decision stability and performance degradation under targeted feature interventions to evaluate the robustness of instruction-conditioned reliance modulation.

\section{Results}
\vspace{-1em}

\subsection{Controlled Modulation of Feature Reliance}
\vspace{-1em}
\begin{figure}[t]
    \centering
    \includegraphics[width=0.85\columnwidth]{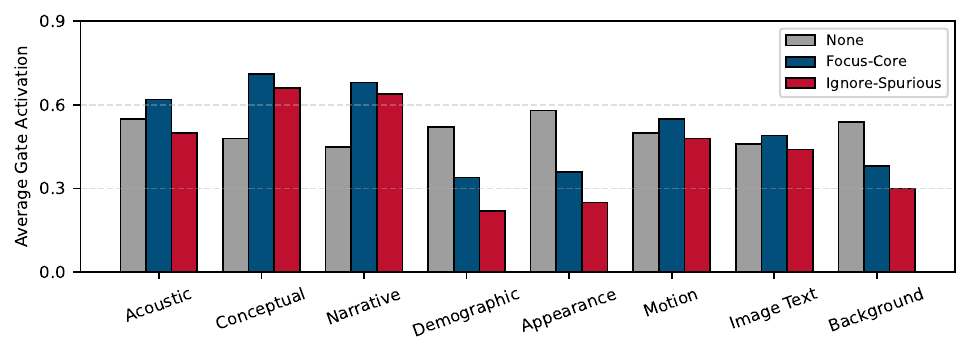}
    \vspace{-1.5em}
    \caption{Instruction-conditioned modulation of taxonomy-aligned feature pathways across datasets. Focus-core instructions consistently amplify semantic and narrative feature groups while suppressing appearance-, demographic-, and background-related pathways. Combined with the corresponding RS analysis, these results indicate that FiLoRA induces structured and functionally meaningful shifts in feature reliance under fixed task semantics.}
    \label{fig:gate_controllability}
    \vspace{-1.5em}
\end{figure}

\begin{wraptable}{r}{0.42\columnwidth}
\vspace{-2.5em}
\caption{
Dataset-wise GMR and RS across taxonomy-aligned feature groups. RS is under focus-core instructions, while GMR measures gate modulation between contrasting
focus/ignore instructions. Higher RS = greater prediction sensitivity to the feature-group gates.
}
\centering

\setlength{\heavyrulewidth}{0.7pt}
\fontsize{7.8}{9.2}\selectfont
\setlength{\tabcolsep}{3.2pt}
\renewcommand{\arraystretch}{0.92}

\begin{tabular}{lcc|cc|cc}
\toprule 
\textbf{Feature}
& \multicolumn{2}{c|}{\textbf{MM-IMDb}} 
& \multicolumn{2}{c|}{\textbf{CREMA-D}} 
& \multicolumn{2}{c}{\textbf{RAVDESS}} \\
\textbf{Group} & GMR & RS & GMR & RS & GMR & RS \\
\midrule
Core     & 0.42 & 0.18 & 0.38 & 0.16 & 0.35 & 0.14 \\
Spurious & 0.47 & 0.07 & 0.41 & 0.08 & 0.39 & 0.06 \\
\bottomrule
\end{tabular}

\label{tab:gmr_rs}
\end{wraptable}
We first evaluate whether natural language instructions induce systematic and semantically aligned shifts in feature reliance, rather than superficial or noisy gate modulation. Figure~\ref{fig:gate_controllability} visualizes average gate activations aggregated over high-level feature families defined by our feature taxonomy. 
Different instruction conditions produce distinct and semantically aligned gate configurations rather than uniform scaling across all feature groups. 
Focus-core instructions amplify gates associated with semantic and narrative pathways, while suppressing appearance-, demographic-, and background-related pathways commonly linked to shortcut reliance.
Table~\ref{tab:gmr_rs} further quantifies these effects using GMR and RS. 
Across all datasets, both core and spurious feature groups exhibit substantial GMR values, indicating that internal gates consistently respond to instruction changes. 
However, RS reveals a clear asymmetry: core-associated groups exhibit substantially higher functional influence on predictions, whereas spurious-associated groups show markedly lower sensitivity despite comparable gate responsiveness.
Dataset-specific feature composition analysis is in Appendix~\ref{appendix:feature_composition}, revealing strong concentration of shortcut-associated cues across modalities and feature groups.
Analysis of dataset-specific reliance structures and feature-label relationships is in Appendix~\ref{appendix:dataset_reliance_profiles}.


\subsection{Functional Redistribution of Prediction Reliance} 
\vspace{-1em}
\begin{wrapfigure}{r}{0.35\columnwidth}
    \vspace{-1.5em}
    \centering
    \includegraphics[width=\linewidth]{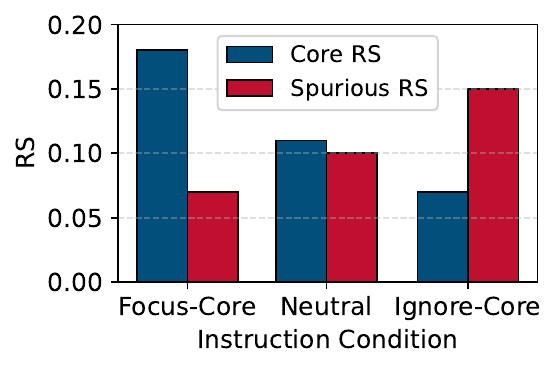}
    \vspace{-2em}

    \caption{Prediction sensitivity to feature-group gate perturbations under different instruction conditions. Focus-core instructions increase sensitivity to core pathways, whereas ignore-core instructions shift sensitivity toward shortcut-associated feature groups.}

    \label{fig:reliance_sensitivity}
    \vspace{-1em}
\end{wrapfigure}

We next evaluate whether instruction-conditioned gate modulation produces measurable functional changes in model behavior, rather than superficial parameter movement.
Figure~\ref{fig:reliance_sensitivity} reports RS, which quantifies how strongly predictions depend on perturbations to different feature-group gates. 
Under focus-core instructions, prediction sensitivity concentrates on semantic and task-relevant pathways, indicating increased dependence on core evidence during prediction.
Conversely, instructions that suppress or de-emphasize core evidence shift sensitivity toward appearance-, demographic-, and shortcut-associated pathways. 
These results demonstrate that instruction-conditioned gating produces systematic redistribution of prediction reliance, rather than merely changing gate magnitudes or inducing representation drift. 
Table~\ref{tab:rs} further quantifies these trends across instruction conditions.
Under focus-core instructions, core-related pathways dominate prediction sensitivity, whereas this relationship reverses under ignore-core instructions. 
Importantly, RS values under neutral instructions remain comparatively balanced, suggesting that the observed redistribution of sensitivity is instruction-induced rather than an artifact of the gating architecture itself.
These findings suggest that FiLoRA enables controllable redistribution of prediction reliance, rather than post hoc analysis of fixed representations. 
Dataset-wise RS results are in Appendix~\ref{appendix:rs_datasetwise}.

\begin{table*}[t]
\centering

\begin{minipage}[t]{0.48\textwidth}
\caption{
RS under different instruction conditions in the instruction-wise evaluation. Higher
RS indicates greater prediction sensitivity to
feature-group gate perturbations
}
\centering
\small
\setlength{\tabcolsep}{4pt}
\renewcommand{\arraystretch}{0.95}

\begin{tabular}{lccc}
\toprule
\textbf{Condition} & \textbf{Core RS} & \textbf{Spurious RS} & \textbf{C/S Ratio} \\
\midrule
Focus-Core  & 0.184 & 0.071 & 2.59 \\
Neutral     & 0.112 & 0.098 & 1.14 \\
Ignore-Core & 0.069 & 0.153 & 0.45 \\
\bottomrule
\end{tabular}

\vspace{-0.4em}
\label{tab:rs}
\end{minipage}
\hfill
\begin{minipage}[t]{0.48\textwidth}
\caption{
Generalization across instruction formulations. 
FiLoRA maintains consistent GMR and RS patterns across template, paraphrased, and compositional instructions.
}
\centering
\small
\setlength{\tabcolsep}{5pt}
\renewcommand{\arraystretch}{0.95}

\begin{tabular}{lccc}
\toprule
\textbf{Type} & \textbf{GMR} & \textbf{Core RS} & \textbf{Spurious RS} \\
\midrule
Template       & 0.42 & 0.18 & 0.07 \\
Paraphrased    & 0.40 & 0.17 & 0.08 \\
Compositional  & 0.39 & 0.16 & 0.09 \\
\bottomrule
\end{tabular}

\label{tab:instruction_generalization}
\end{minipage}
\vspace{-1.2em}
\end{table*}

\subsection{Generalization to Natural and Compositional Instructions}
\vspace{-0.7em}
We next evaluate whether FiLoRA generalizes beyond fixed instruction templates to more natural and compositional language forms.
Table~\ref{tab:instruction_generalization} shows that FiLoRA maintains consistent controllability behavior across template, paraphrased, and compositional instructions. Although minor variation is observed in absolute values, the relative ordering between core and spurious reliance remains stable.
These results suggest that FiLoRA responds to the semantic direction of instructions rather than memorizing fixed lexical templates.

\begin{figure}[t]
    \centering
    \includegraphics[width=\columnwidth]{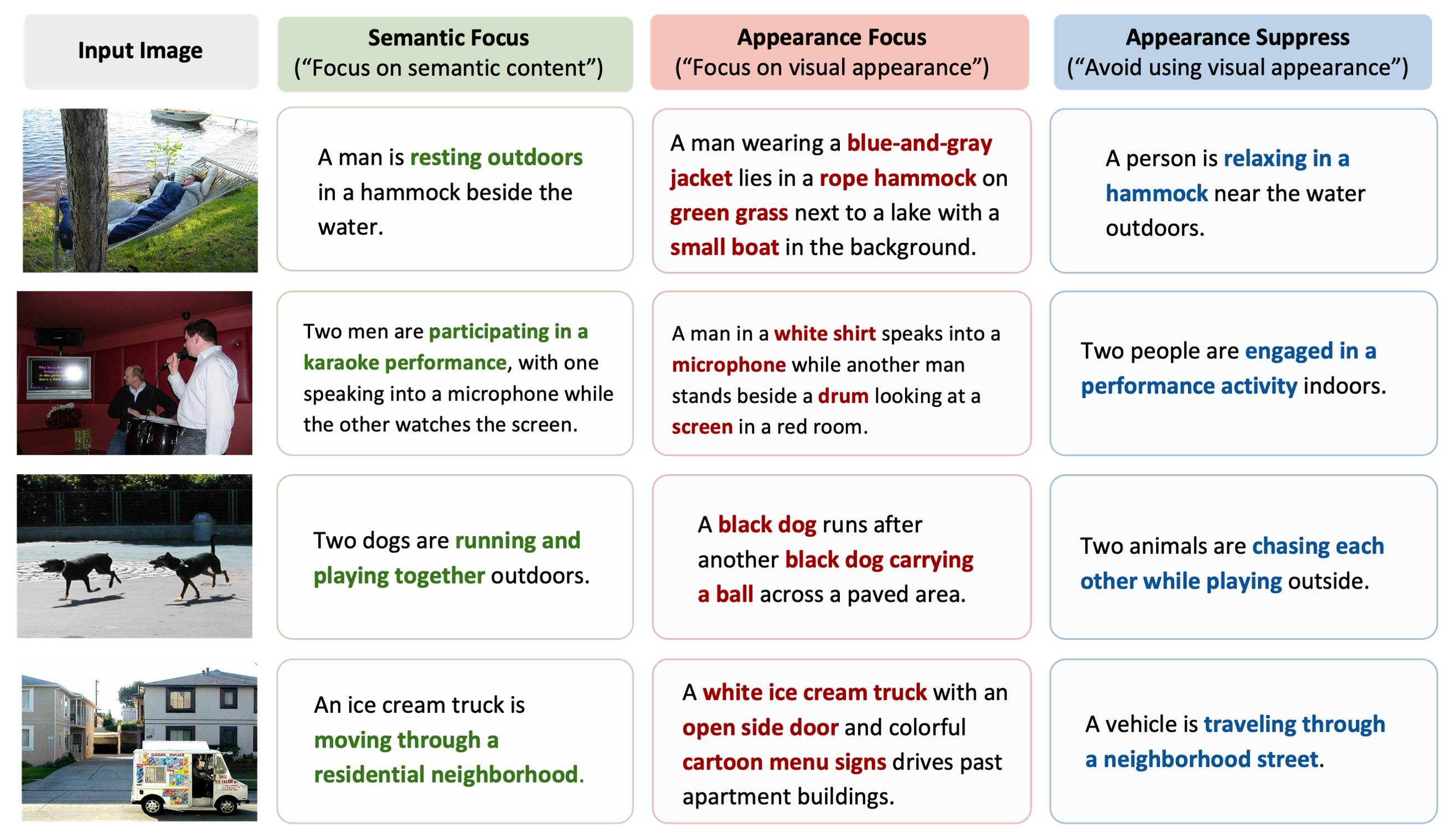}
    \vspace{-1em}
    \caption{
Example captions generated under different instruction conditions on Flickr30K. 
    FiLoRA produces instruction-aligned changes in caption content: semantic-focused instructions emphasize high-level scene meaning and relationships, whereas appearance-focused instructions shift caption emphasis toward visual attributes such as color, clothing, and background appearance.
    }
    \label{fig:flickr_caption_examples}
    \vspace{-1.3em}
\end{figure}

\vspace{-0.7em}
\subsection{Generative Evaluation on Image Captioning}
\vspace{-0.7em}

\begin{wraptable}{r}{0.52\columnwidth}
\vspace{-1.8em}
\caption{
Generative evaluation on Flickr30K. 
FiLoRA preserves competitive caption quality while enabling controllable shifts in feature reliance.
}

\centering
\small
\setlength{\tabcolsep}{3.5pt}
\renewcommand{\arraystretch}{0.92}

\resizebox{\linewidth}{!}{
\begin{tabular}{lccc}
\toprule
\textbf{Instruction Condition} & \textbf{CIDEr} & \textbf{CLIPScore} & \textbf{GMR} \\
\midrule
Semantic Focus      & 112.4 & 0.823 & 0.41 \\
Appearance Focus    & 109.7 & 0.817 & 0.39 \\
Appearance Suppress & 113.1 & 0.826 & 0.43 \\
\bottomrule
\end{tabular}
}

\label{tab:flickr_captioning}
\vspace{-1em}

\end{wraptable}

We evaluate whether instruction-conditioned reliance modulation extends beyond controlled classification settings to open-ended multimodal generation. We adopt FiLoRA on Flickr30K image captioning using instruction conditions that emphasize or suppress different visual and semantic feature groups during caption generation. Unlike classification tasks, caption generation requires continuous integration of multimodal evidence across decoding steps, providing a substantially more challenging setting for controllable reliance modulation.
Figure~\ref{fig:flickr_caption_examples} shows representative examples generated under different instruction conditions. 
When instructed to focus on semantic content, FiLoRA generates captions emphasizing interactions, activities, and contextual relationships between entities. 
In contrast, appearance-focused instructions increase attention to low-level visual attributes such as color, clothing, and background appearance.
Table~\ref{tab:flickr_captioning} shows that FiLoRA maintains competitive caption quality across diverse instruction conditions while preserving substantial gate modulation behavior. 
Semantic-focused instructions prioritize scene meaning and contextual relationships, whereas appearance-focused instructions emphasize low-level visual details without changing the underlying captioning task.
These results suggest that FiLoRA generalizes beyond controlled classification settings and supports structured control over feature-level reliance in open-ended multimodal generation. 
Additional examples and the full instruction taxonomy evaluation table, including conceptual, behavioral, contextual, and compositional instruction variants, are provided in Appendix~\ref{appendix:caption_taxonomy_examples}.

\subsection{Sensitivity to Feature Grouping Strategy}

\begin{wraptable}{r}{0.40\columnwidth}
\vspace{-1.5em}
\caption{
Sensitivity to feature grouping strategy.
\textbf{Semantically aligned grouping (ours)} achieves stronger instruction-conditioned gate modulation and greater separation between core- and spurious-associated prediction sensitivity. C/S denotes the ratio of Core to Spurious RS.
}
\centering
\small
\setlength{\tabcolsep}{2.5pt}
\renewcommand{\arraystretch}{0.92}
\begin{tabular}{lcc}
\toprule
\textbf{Grouping Strategy} & \textbf{C/S RS Ratio} & \textbf{GMR} \\
\midrule
\textbf{Semantic Grouping} & 2.47 & 0.42 \\
Coarse Grouping & 1.53 & 0.31 \\
Random Grouping & 1.08 & 0.17 \\
\bottomrule
\end{tabular}
\label{tab:grouping_sensitivity}
\vspace{-0.8em}
\end{wraptable}

We next examine how the construction of trainable LoRA
adaptation groups affects instruction-conditioned
controllability. We compare three strategies:
(i) semantically aligned grouping (ours), in which
adaptation groups are associated with interpretable,
task-relevant evidence categories;
(ii) coarse architectural grouping, in which groups
are organized according to model components without
explicit semantic alignment; and
(iii) random grouping, in which groups are assigned
without reference to semantic evidence categories.
Table~\ref{tab:grouping_sensitivity} shows that
semantically aligned grouping achieves the strongest
controllability, with higher GMR and greater separation
between core- and spurious-associated prediction
sensitivity. Coarse architectural grouping yields
intermediate results, while random grouping exhibits
the weakest performance.
These findings suggest that semantically meaningful
organization of LoRA adaptation groups provides a
useful inductive bias for instruction-conditioned
feature reliance modulation, rather than being a
strict requirement of the FiLoRA framework.
Additional analysis of proxy supervision quality
is provided in Appendix~\ref{appendix:proxy_analysis}.

\subsection{Robustness Under Spurious Feature Interventions}

\begin{wrapfigure}{r}{0.6\columnwidth}
\vspace{-1.5em}
    \centering

    \subfigure{
        \includegraphics[width=0.45\linewidth]{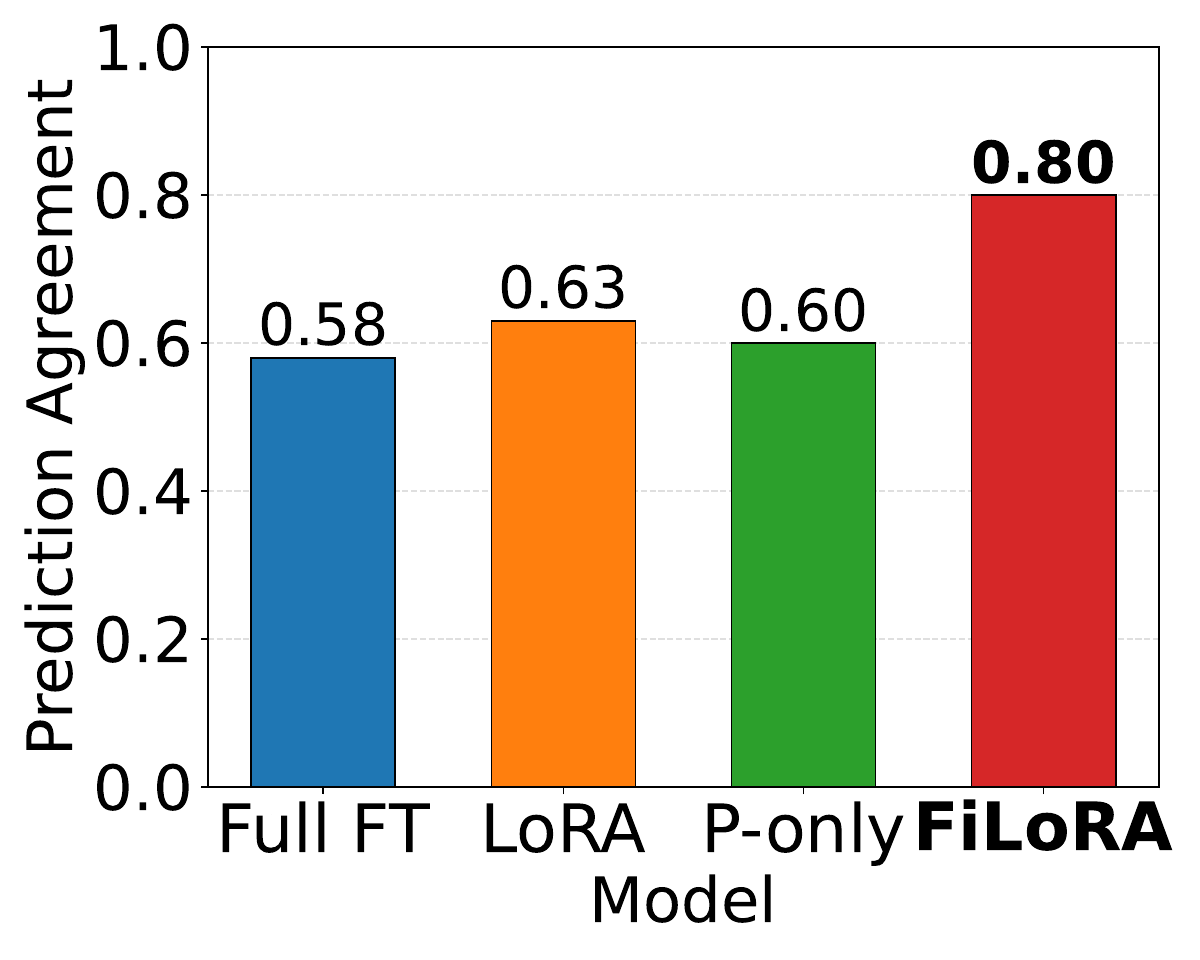}
        \label{fig:decision_stability_spurious}
    }
    \hfill
    \subfigure{
        \includegraphics[width=0.45\linewidth]{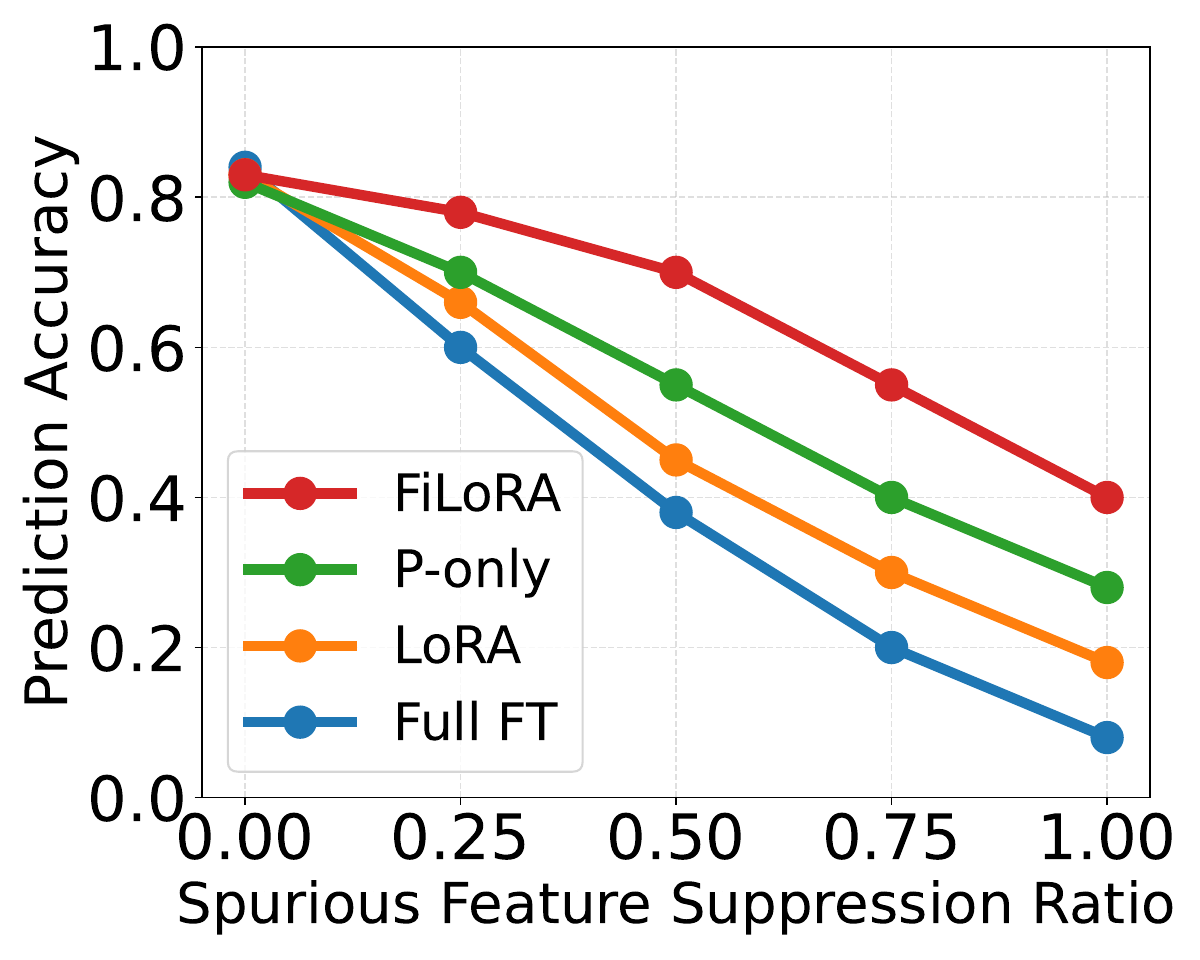}
        \label{fig:spurious_degradation_curve}
    }

    \vspace{-0.5em}
    \caption{Decision stability under spurious feature removal (left) and performance degradation under progressively increasing spurious feature suppression (right). 
Compared to full fine-tuning, LoRA, and prompt-only baselines, FiLoRA preserves higher prediction stability and degrades more gradually under shortcut-associated feature interventions. Detailed definitions of intervention protocols and stability metrics are provided in Appendix~\ref{appendix:robustness_dataset}.}
    
    \label{fig:robustness}
    \vspace{-1em}
\end{wrapfigure}

We next compare FiLoRA against adaptation and prompting baselines under targeted spurious feature interventions.
Figure~\ref{fig:robustness}(left) reports prediction agreement before and after spurious feature removal. Full fine-tuning, LoRA, and prompt-only baselines exhibit substantial instability, indicating strong dependence on shortcut-associated cues. In contrast, FiLoRA preserves substantially higher decision stability.
Figure~\ref{fig:robustness}(right) further shows performance degradation under progressively increasing spurious feature suppression. Baseline models degrade sharply even under mild suppression, whereas FiLoRA exhibits a substantially more gradual decline.
These results suggest that FiLoRA improves robustness through explicit regulation of internal feature reliance rather than output-level prompting or standard parameter-efficient adaptation alone. Additional cross-architecture results are provided in Appendix~\ref{appendix:architecture_generalization}.


\section{Conclusion}
We introduced FiLoRA, a parameter-efficient framework for instruction-conditioned modulation of feature reliance in multimodal models.
By combining grouped LoRA adaptation with instruction-conditioned gating, FiLoRA enables natural language instructions to regulate the contributions of feature-aligned adaptation pathways while preserving a common prediction task and label space.
Unlike static adaptation methods, FiLoRA provides instruction-specific control over multiple evidence-associated pathways without modifying the pretrained backbone or requiring inference-time retraining.
Across classification and generative multimodal settings, our experiments demonstrate systematic changes in gate activation and prediction sensitivity, generalization across diverse instruction formulations, and improved robustness under targeted spurious feature interventions.
Importantly, these results show that instruction-conditioned modulation extends beyond changes in gate magnitudes, producing measurable differences in how adaptation pathways influence model predictions.
Furthermore, consistent reliance patterns across heterogeneous multimodal backbones suggest that the framework is applicable across different architectures and model scales.
Together, these findings highlight instruction-conditioned parameter adaptation as a practical approach to influencing how multimodal models use information, extending instruction following beyond output-level behavior toward structured control of feature reliance.

\section*{Acknowledgement}
This research was supported by 
the Institute of Information \& communications Technology Planning \& Evaluation (IITP) grant funded by the Korea government(MSIT) (No.RS-2025-02217259, Development of self-evolving AI bias detection-correction-explain platform based on international multidisciplinary governance), (No.RS-2026-25519206, Development of human-like intelligent generative agents based on interactive multimodal reverse prompting), (RS2024-00395401, Development of VFX creation and combination using generative AI)
and the Korea Planning \& Evaluation Institute of Industrial Technology (KEIT) funded by the Ministry of Trade, Industry and Energy (No.RS-202525458052, Development of Core Technologies for Manufacturing Foundation Models), funding from the Google Award for Inclusion Research Program (G222897), and the Korea Institute of Science and Technology (KIST) Institutional Program (Project no. 26E0121 and 26V0032).

\nocite{langley00}
\bibliography{example_paper}
\bibliographystyle{plainnat}

\newpage
\appendix
\onecolumn
\section{Technical Reference}
\label{sec:appen_tech_ref}
\paragraph{Datasets} All dataset links provided refer to the official sources released by the original authors or dataset maintainers, ensuring authenticity and traceability. We intentionally selected official resources to support reproducibility and transparency of the experimental setup. For MM-IMDb, we use a reconstructed subset of the dataset as a deliberate design choice to better isolate the effect of instruction-conditioned reliance, rather than to maximize task performance.

\begin{enumerate}
    \item Tiny MM-IMDb: \url{https://www.kaggle.com/datasets/gabrieltardochi/tiny-mm-imdb}
    \item CREMA-D: \url{https://gitlab.com/cs-cooper-lab/crema-d-mirror}
    \item RAVDESS: \url{https://zenodo.org/records/1188976\#.Xpaa3i-caAP}
\end{enumerate}

\paragraph{Models}
We use Qwen2.5-VL-7B-Instruct(\url{https://huggingface.co/Qwen/Qwen2.5-VL-7B-Instruct}), Qwen2.5-VL-32B-Instruct(\url{https://huggingface.co/Qwen/Qwen2.5-VL-32B-Instruct}), and Gemma-3n-E4B-it(\url{https://huggingface.co/google/gemma-3n-E4B-it}) for the text--image modality setting, and Qwen2.5-Omni-7B(\url{https://huggingface.co/Qwen/Qwen2.5-Omni-7B}), AV-HuBERT(\url{https://huggingface.co/vumichien/AV-HuBERT}), and AudioCLIP(\url{https://github.com/AndreyGuzhov/AudioCLIP}) for the audio--video modality setting, with all models obtained exclusively from their official releases to ensure authenticity and reproducibility. 

The Qwen2.5-VL family provides strong general-purpose vision--language capabilities, Gemma-3n-E4B-it serves as a lightweight instruction-tuned multimodal baseline. Also, Qwen2.5-Omni-7B supports unified multimodal understanding across audio, vision, and language, AV-HuBERT offers robust self-supervised audio--visual speech representations, and AudioCLIP enables aligned representation learning across audio, vision and text, collectively allowing us to evaluate instruction-conditioned reliance across diverse multimodal architectures and learning paradigms. A detailed description of the experimental protocols involving these models, as well as the corresponding results, is provided in Appendix~\ref{appendix:architecture_generalization}.

\section{Dataset-Intrinsic Feature Composition Analysis}
\label{appendix:feature_composition}
\begin{figure}[p]
    \centering
    \includegraphics[width=\textwidth]{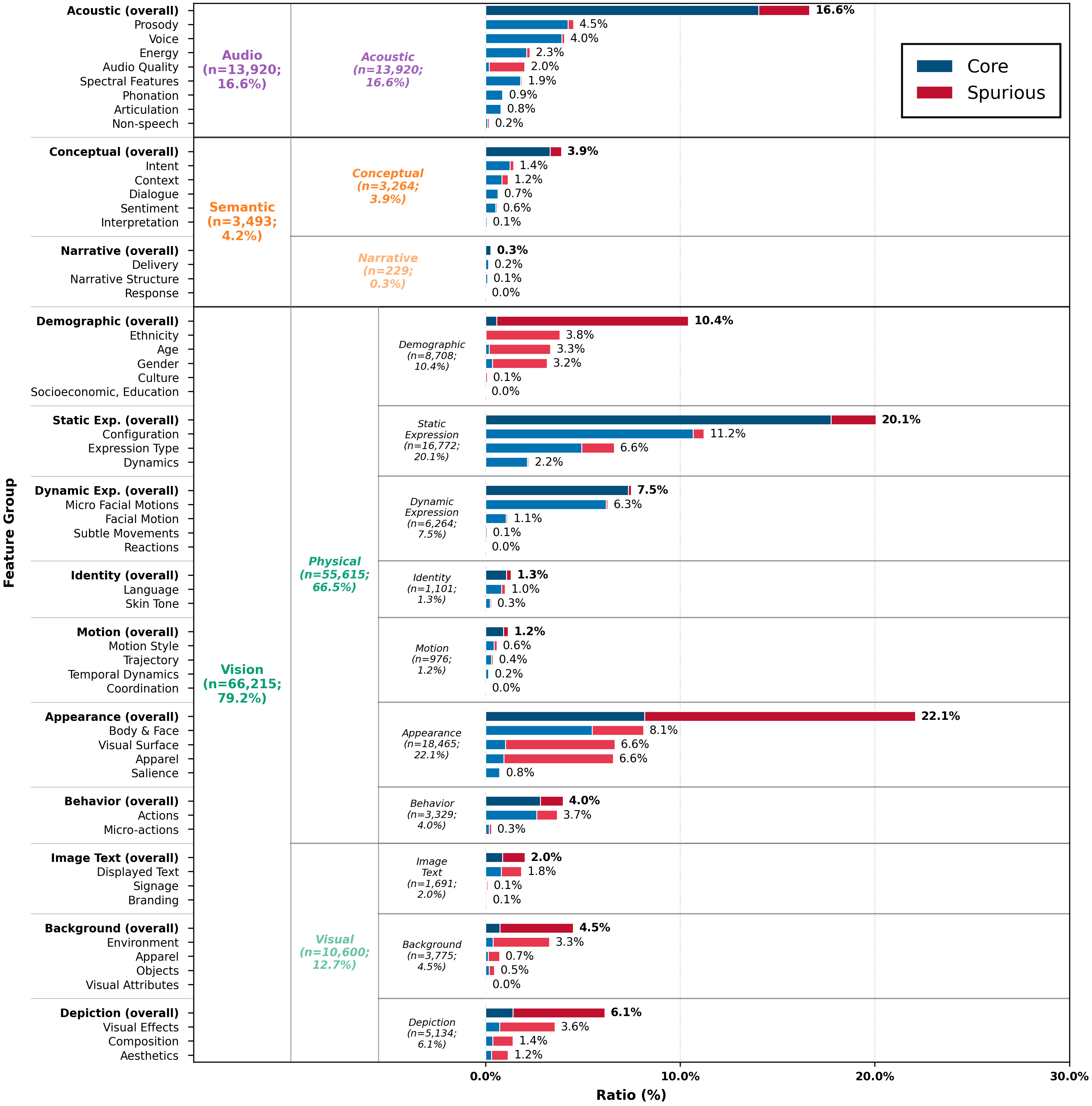}
    \caption{Aggregated core and spurious feature composition across all datasets. Visual appearance, background, and static expression features account for a large fraction of spurious annotations, whereas semantic and narrative features are predominantly core-aligned. This aggregation highlights recurring structural patterns that make shortcut learning advantageous in multimodal settings.}
    \label{fig:feature_composition_total}
\end{figure}

\begin{figure}[p]
    \centering
    \includegraphics[width=\textwidth]{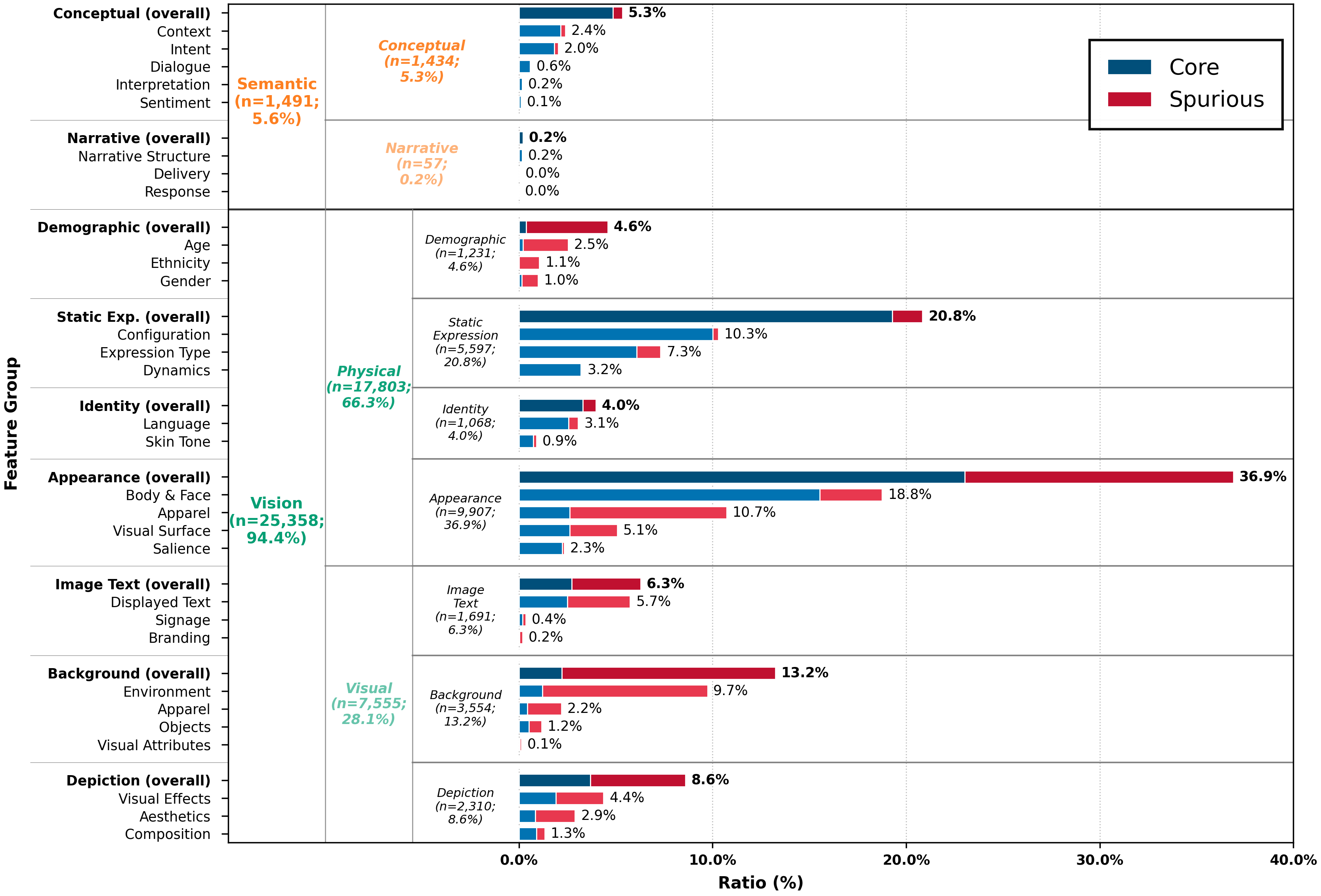}
    \caption{Dataset-intrinsic composition of core and spurious feature annotations for MM-IMDb. The dataset exhibits strong visual dominance, with appearance-related features accounting for a large share of annotations. Compared to the other datasets, MM-IMDb contains a higher proportion of spurious visual features, indicating favorable conditions for visual shortcut learning.}
    \label{fig:feature_composition_mmimdb}
\end{figure}

\begin{figure}[p]
    \centering
    \includegraphics[width=\textwidth]{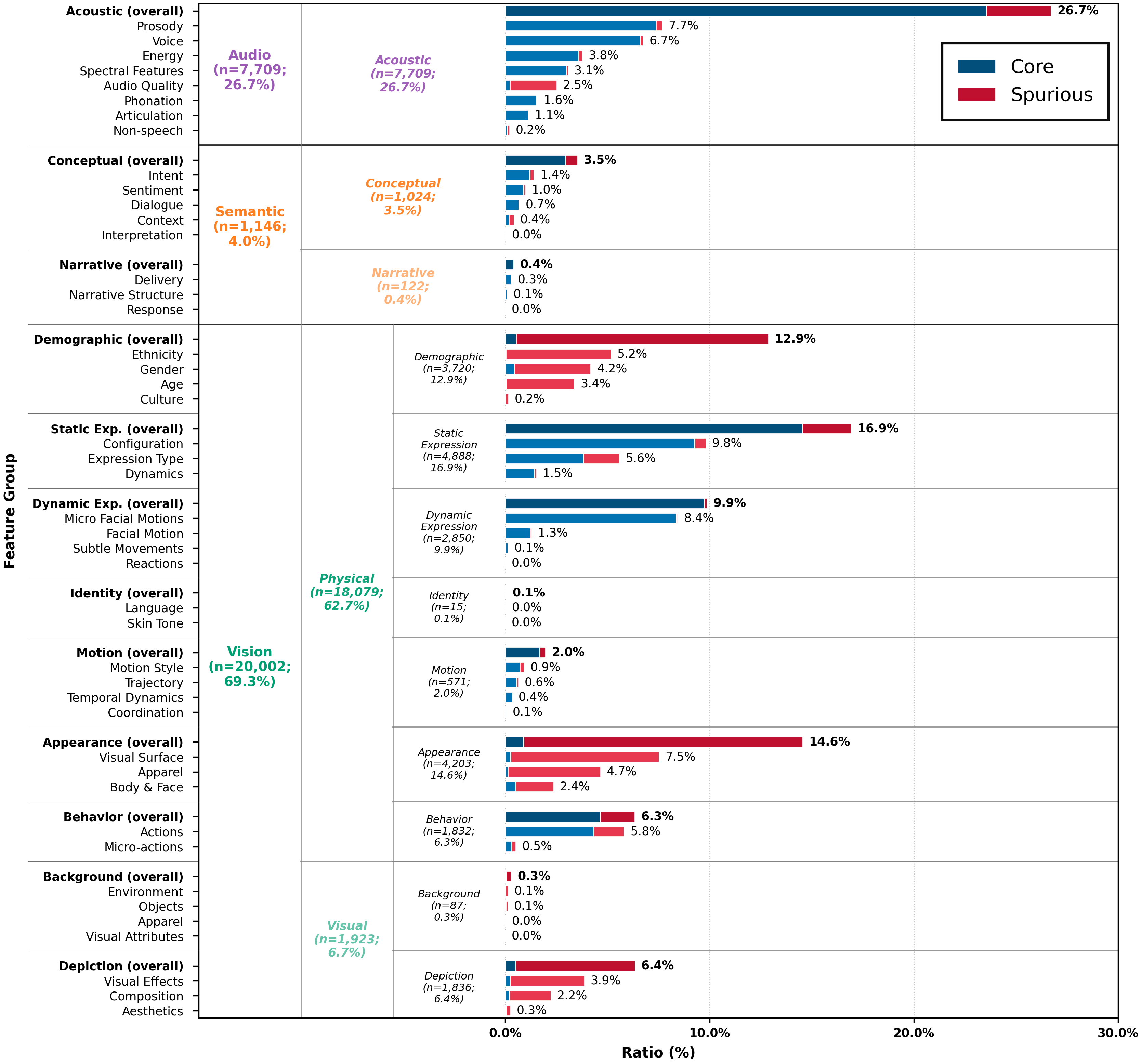}
    \caption{Dataset-intrinsic composition of core and spurious feature annotations for CREMA-D. Emotion-relevant expression features are predominantly core-aligned, while demographic cues contribute a non-negligible share of spurious annotations. Overall modality distributions and core--spurious patterns are similar to those observed in RAVDESS.}
    \label{fig:feature_composition_cremad}
\end{figure}

\begin{figure}[p]
    \centering
    \includegraphics[width=\textwidth]{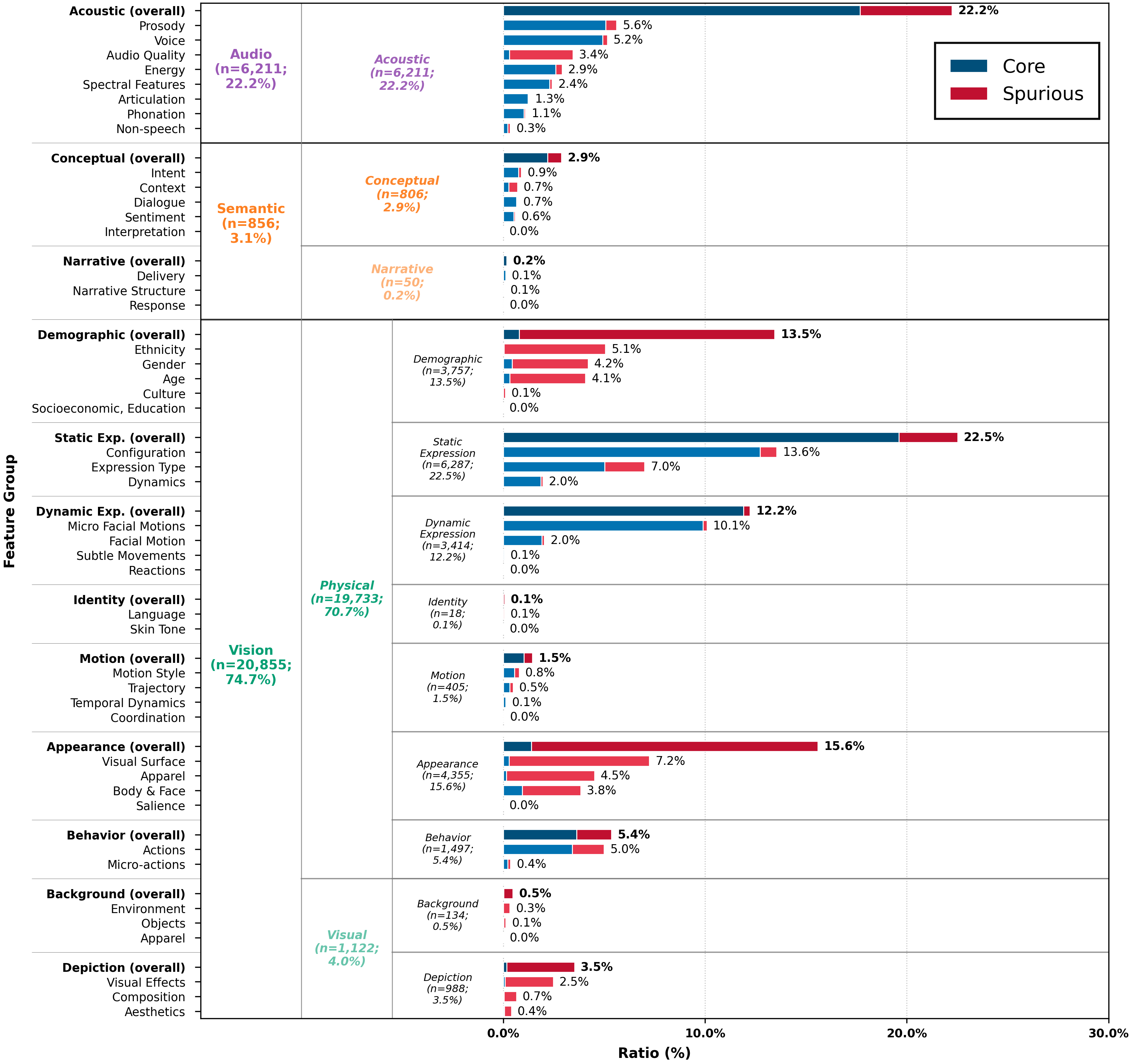}
     \caption{Dataset-intrinsic composition of core and spurious feature annotations for RAVDESS, organized by modality and feature group. Acoustic and expression-related features dominate the annotation space and are predominantly core-aligned, while demographic and background cues appear less frequently.}
    \label{fig:feature_composition_ravdess}
\end{figure}

We analyze the intrinsic composition of core and spurious features across datasets, independent of any model predictions.
For each dataset, we aggregate annotated feature tags according to a hierarchical feature taxonomy and report their relative proportions by modality and feature group. Figure \ref{fig:feature_composition_total} shows the overall feature composition across all datasets, while Figures \ref{fig:feature_composition_mmimdb}, \ref{fig:feature_composition_cremad}, and \ref{fig:feature_composition_ravdess} present the detailed breakdowns for MM-IMDb, CREMA-D, and RAVDESS, respectively.

Across all datasets, core and spurious features are not evenly distributed but instead exhibit pronounced skew toward one role within most feature groups.
In particular, demographic- and background-related features consistently show a high proportion of spurious annotations, indicating that these cues are structurally well-suited for shortcut learning.
In contrast, semantic, narrative, and expression-related features tend to be predominantly core-aligned.

We observe substantial dataset-specific differences in how these feature groups are distributed.
MM-IMDb exhibits extreme visual dominance, with appearance-related features accounting for the largest share of annotations.
This dataset also contains a higher overall proportion of spurious visual features compared to the other datasets, suggesting that visual shortcuts are especially accessible.
By contrast, CREMA-D and RAVDESS display more balanced modality distributions, with emotion-relevant features such as static and dynamic expressions exhibiting high core ratios.

Cross-dataset comparisons further reveal modality-specific structure.
The image-based MM-IMDb dataset is dominated by appearance-related features, whereas the video-based datasets (CREMA-D and RAVDESS) are dominated by static expression features.
Background and identity-related cues (e.g., skin tone and language) are rare in CREMA-D and RAVDESS, consistent with their controlled recording environments, but appear more frequently in MM-IMDb.

These dataset-intrinsic statistics motivate our focus on instruction-conditioned reliance control.
When spurious cues are concentrated within specific modalities or feature groups, shortcut learning becomes structurally advantageous unless reliance is explicitly regulated.

\begin{figure}[t]
    \centering
    \subfigure[MM-IMDb]{
        \includegraphics[width=0.31\linewidth]{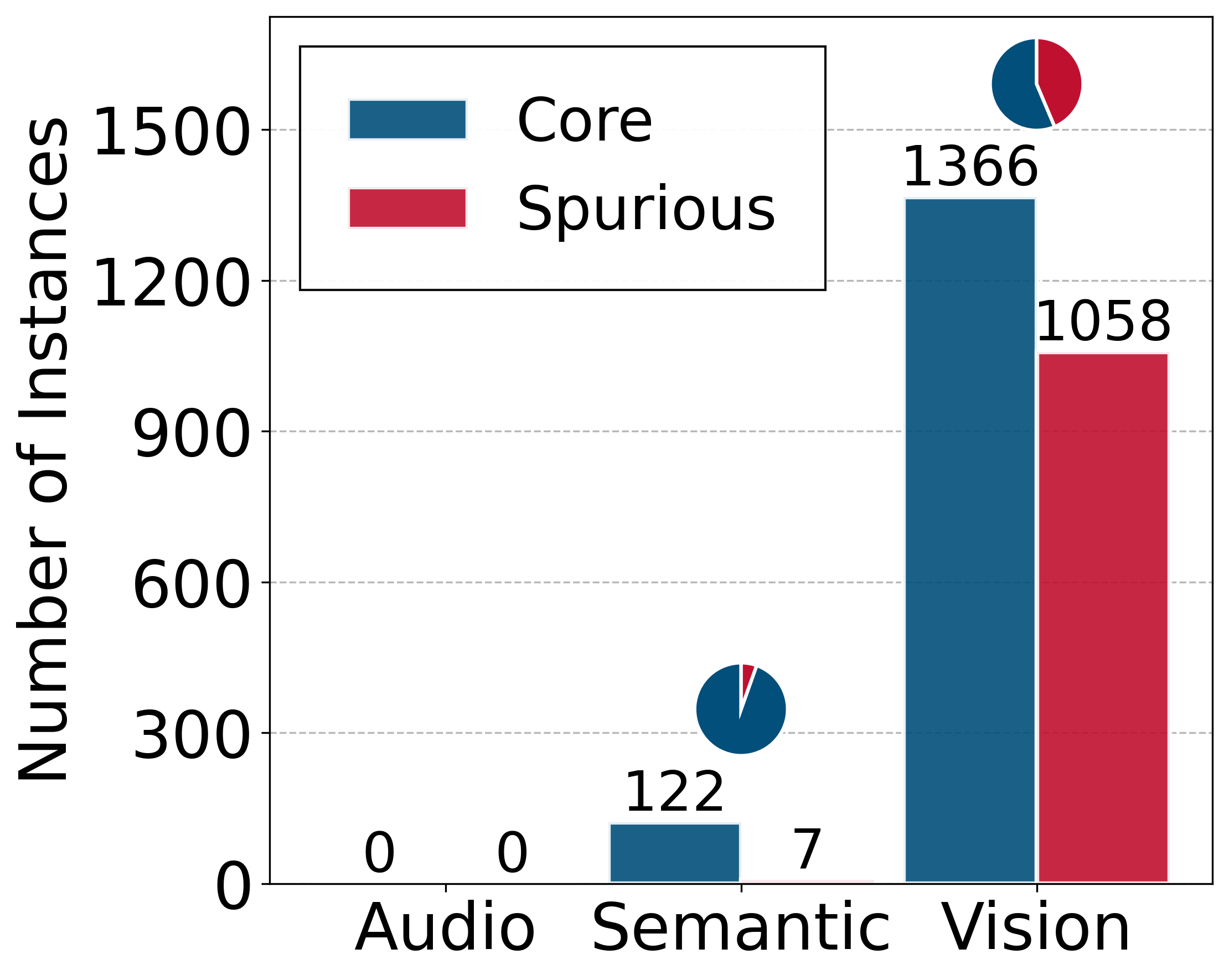}
    }
    \subfigure[CREMA-D]{
        \includegraphics[width=0.31\linewidth]{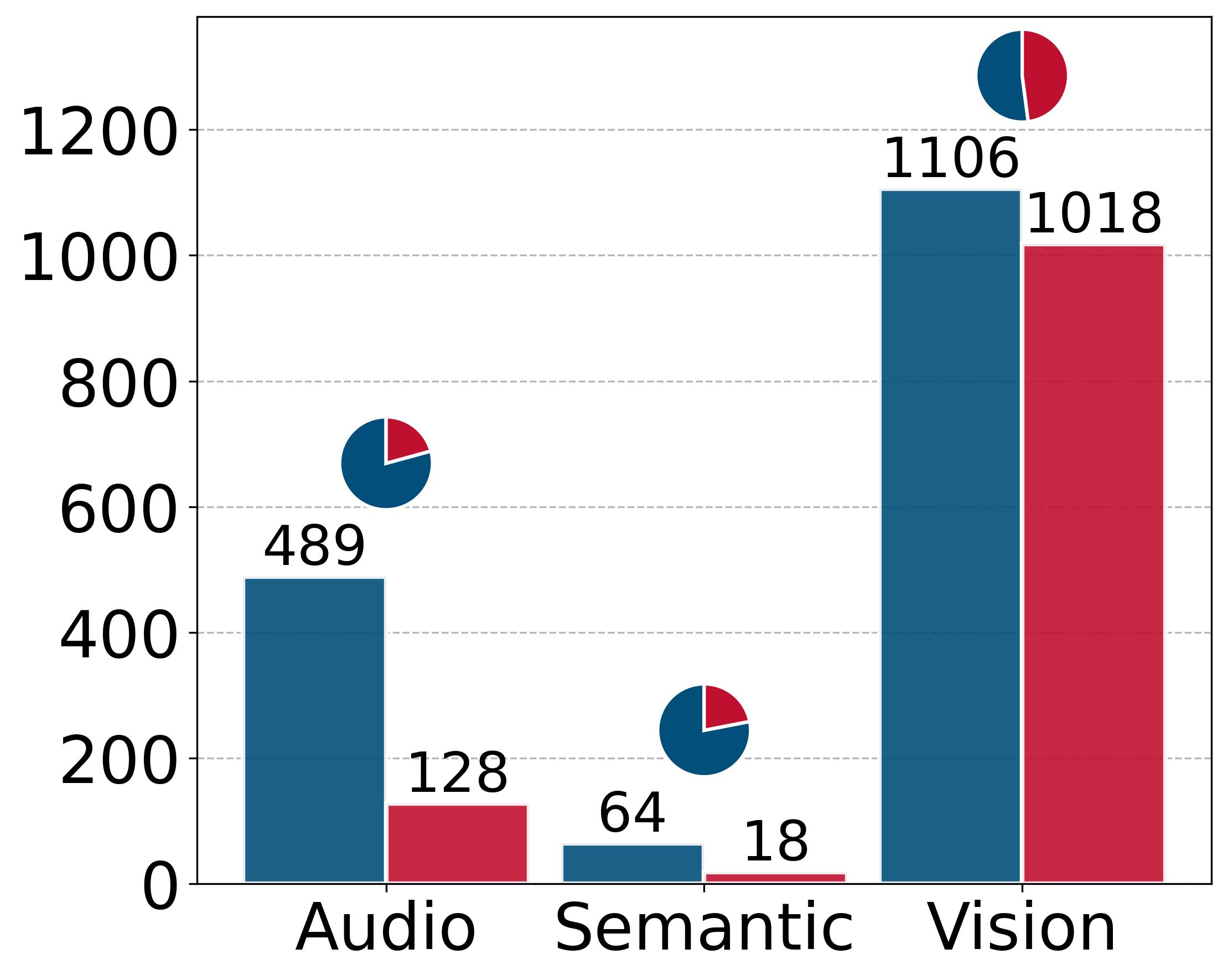}
    }
    \subfigure[RAVDESS]{
        \includegraphics[width=0.31\linewidth]{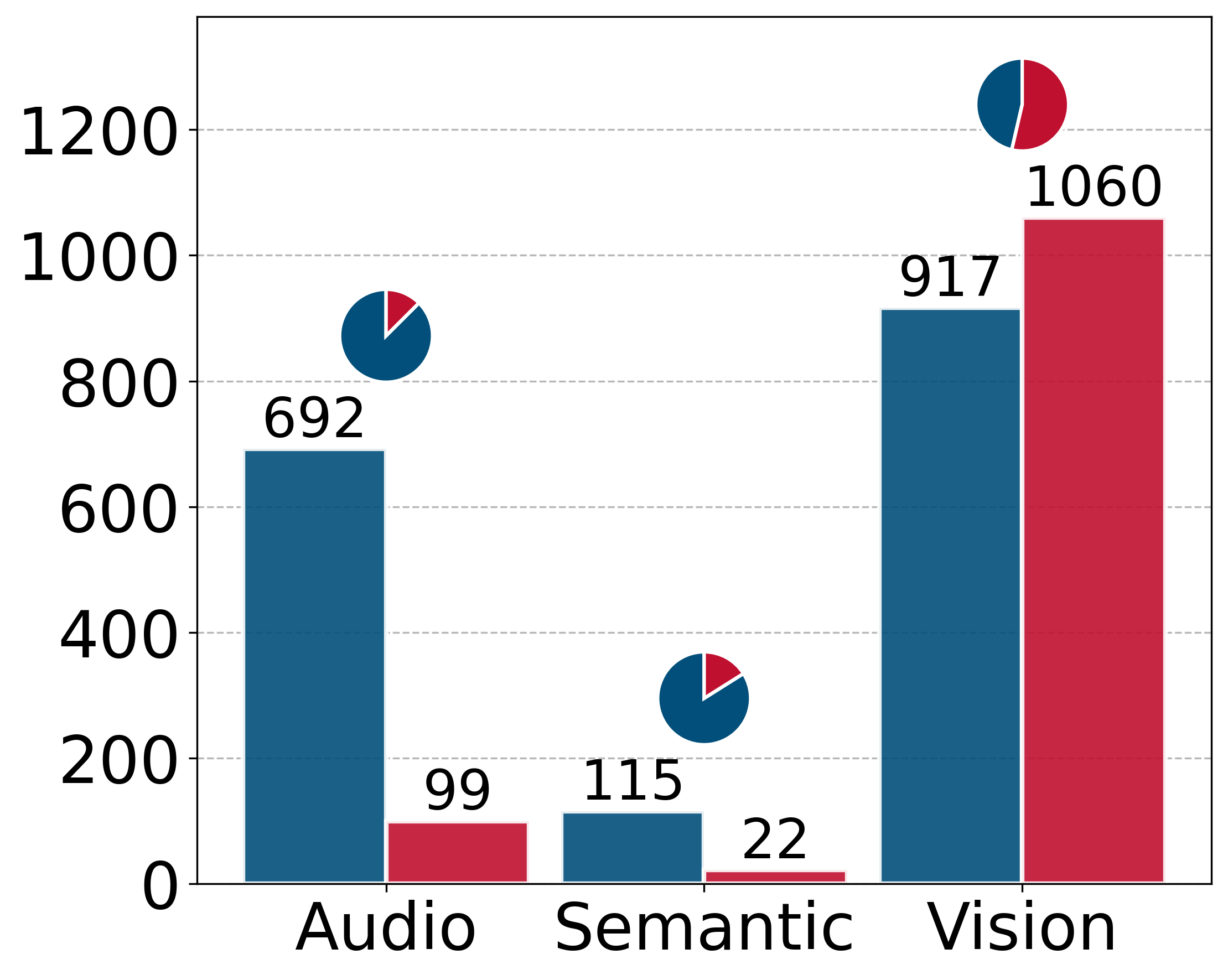}
    }
    \caption{Modality-wise counts of annotated core and spurious features for MM-IMDb, CREMA-D, and RAVDESS. Bars indicate the total number of feature annotations per modality, while inset pies show the relative core-spurious composition within each modality. Vision accounts for the largest volume of annotations across datasets, highlighting its potential as a dominant source of shortcut cues.}
    \vspace{-1em}
    \label{fig:modality_wise_count_core_spurious}
\end{figure}

\FloatBarrier

\vspace{-0.5em}
\section{Modality-Level Distribution of Core and Spurious Features}
\label{appendix:modality_distribution}

This section reports the absolute number of annotated core and spurious features by modality for each dataset, as shown in Figure~\ref{fig:modality_wise_count_core_spurious}.
Unlike ratio-based analyses, this view highlights the raw volume of available cues within each modality, which determines the size of the shortcut search space prior to any model training.
Across MM-IMDb, CREMA-D, and RAVDESS, vision consistently accounts for the largest number of feature annotations, encompassing both core and spurious cues.
This dominance indicates that visual modalities provide a particularly rich and diverse set of signals, making shortcut learning structurally more accessible than in audio or semantic modalities.
While the overall prominence of vision is consistent, the relative balance between core and spurious features within vision varies across datasets.
MM-IMDb and CREMA-D exhibit a higher proportion of core-aligned visual features, whereas RAVDESS shows a comparatively larger share of spurious visual annotations.
In contrast, audio and semantic modalities are consistently core-dominant across all datasets, suggesting that task-relevant information is more uniformly represented in these modalities.
Differences in the semantic modality further reflect dataset characteristics.
In MM-IMDb, semantic features are predominantly core-aligned, consistent with the relevance of plot descriptions to genre labels.
By contrast, in CREMA-D and RAVDESS, semantic annotations show a higher spurious proportion, reflecting the weaker association between spoken content and emotion labels.
Overall, this modality-level analysis complements the feature group breakdown presented in Appendix~\ref{appendix:feature_composition}.
Together, these analyses characterize where shortcut-prone cues are concentrated in the data and motivate the need for explicit, instruction-conditioned regulation of modality reliance.

%

\begin{figure}
    \centering
    \includegraphics[width=1\linewidth]{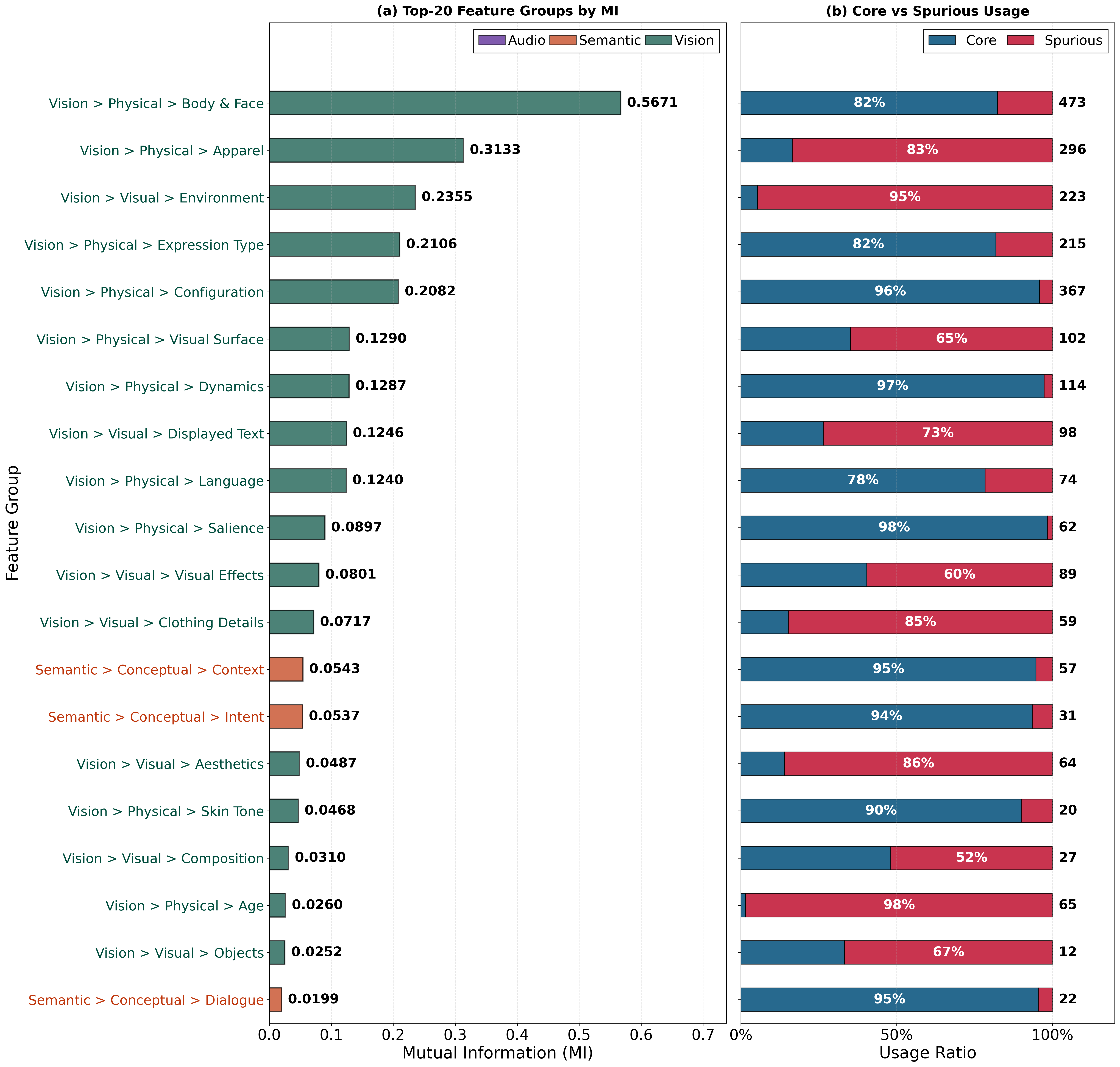}
    \caption{Top-20 feature groups ranked by mutual information (MI) with target labels for the MM-IMDb dataset. Panel (a) shows MI values by feature group, and panel (b) reports the corresponding core versus spurious usage ratios. Highly predictive feature groups are not necessarily core-dominant, revealing dataset-specific shortcut structures.}
    \label{fig:label_association_mmimdb}
\end{figure}

\begin{figure}
    \centering
    \includegraphics[width=1\linewidth]{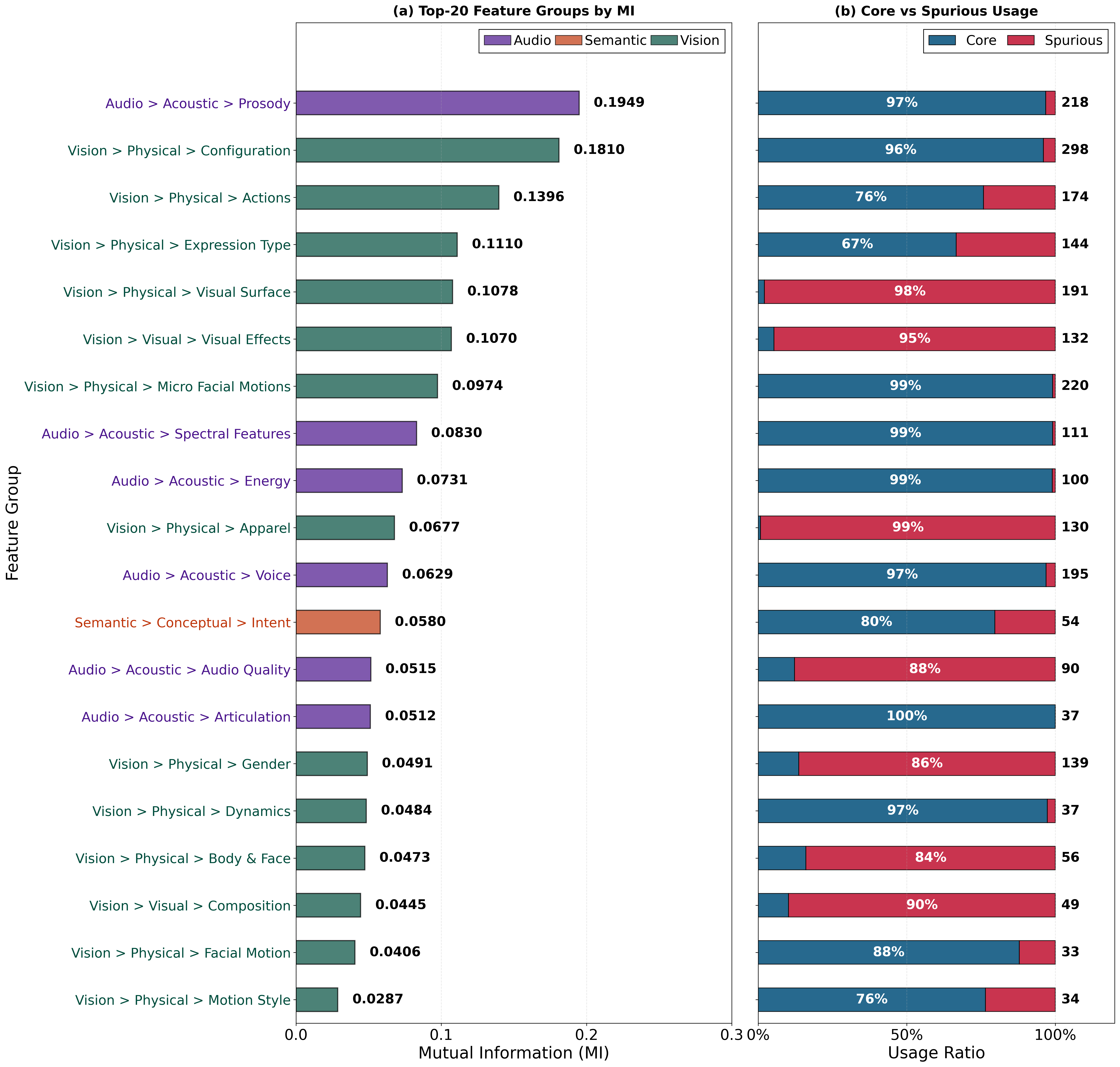}
    \caption{Top-20 feature groups ranked by mutual information (MI) with target labels for the CREMA-D dataset. Panel (a) shows MI values by feature group, and panel (b) reports the corresponding core versus spurious usage ratios. Highly predictive feature groups are not necessarily core-dominant, revealing dataset-specific shortcut structures.}   
    \label{fig:label_association_cremad}
\end{figure}

\begin{figure}
    \centering
    \includegraphics[width=1\linewidth]{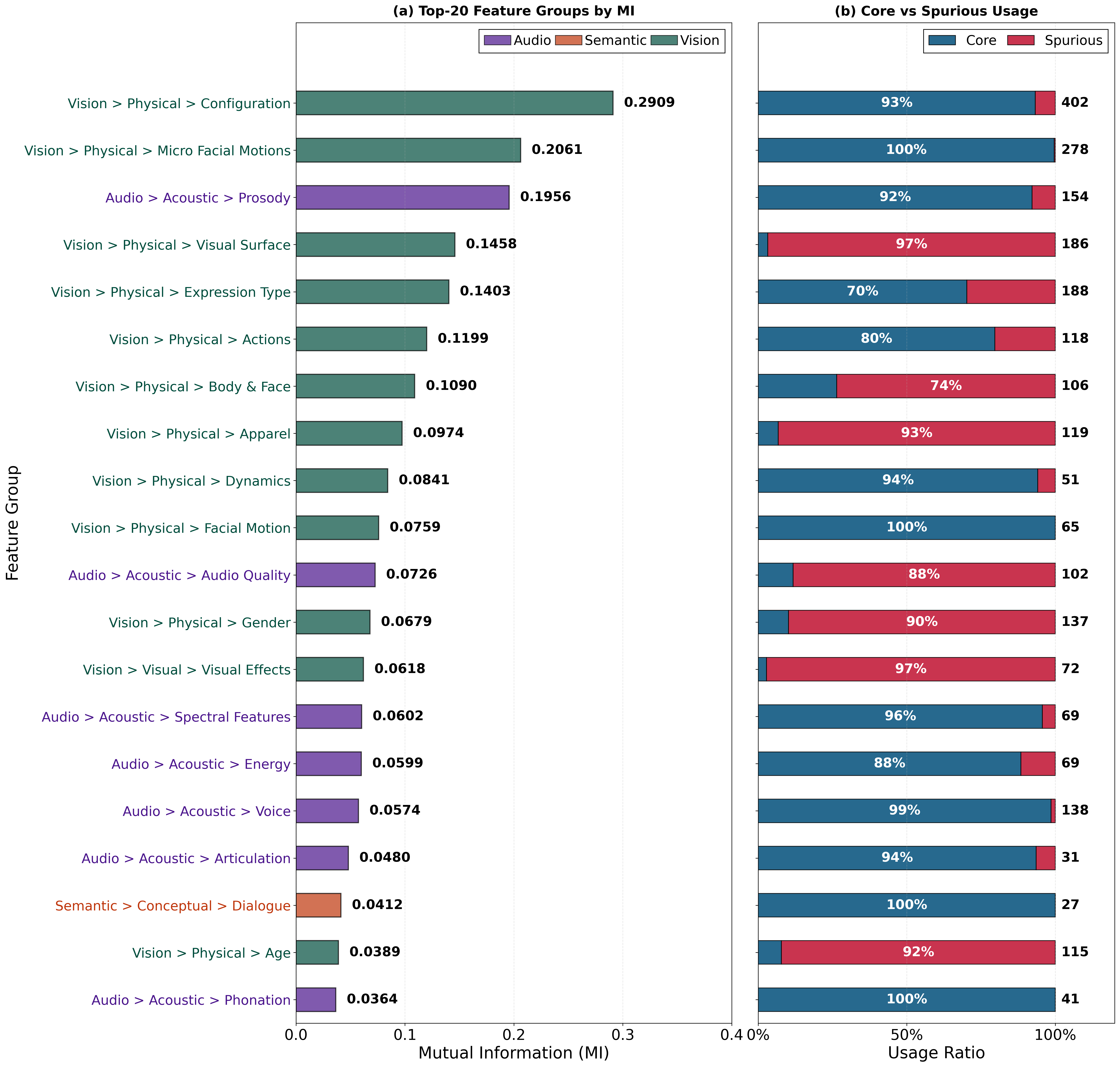}
    \caption{Top-20 feature groups ranked by mutual information (MI) with target labels for the RAVDESS dataset. Panel (a) shows MI values by feature group, and panel (b) reports the corresponding core versus spurious usage ratios. Highly predictive feature groups are not necessarily core-dominant, revealing dataset-specific shortcut structures.}
    \label{fig:label_association_ravdess}
\end{figure}

%
%
%
\vspace{-0.8em}
\section{Label-Feature Associations and Core-Spurious Misalignment}
\label{appendix:label_feature_mi}
\begin{figure}[t]
    \centering
    \includegraphics[width=\columnwidth]{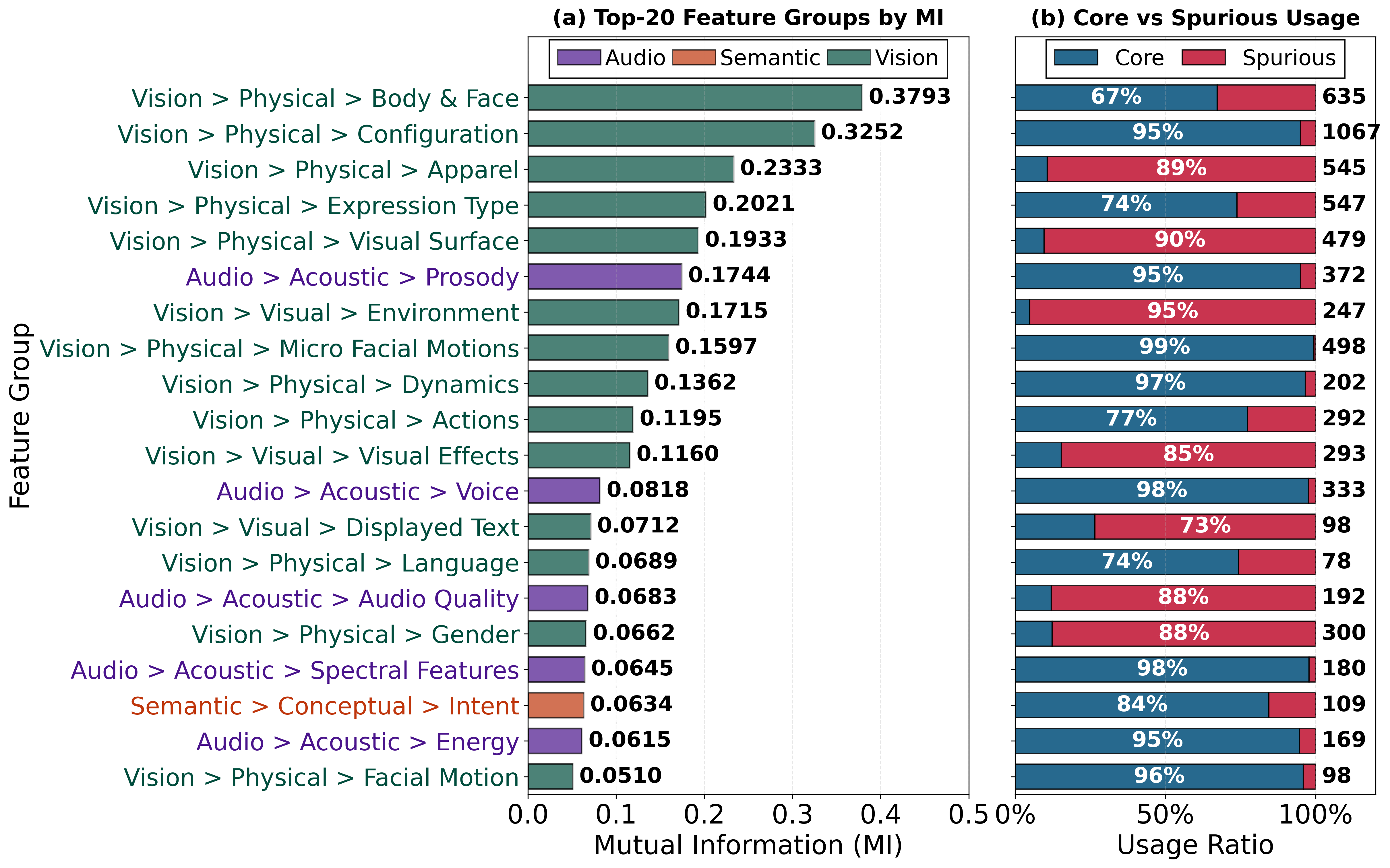}
    \vspace{-1.5em}
\caption{Top-20 feature groups ranked by mutual information with labels.
    Several highly label-associated features are spurious-dominant,
    highlighting that correlation-driven learning alone cannot distinguish task-relevant signals from shortcut cues.}
    \label{fig:top10_mi}
    \vspace{-1em}
\end{figure}

This appendix examines which feature groups are most strongly associated with target labels prior to any model intervention. 
For each dataset, we compute the mutual information (MI) between feature group annotations and ground-truth labels, and rank feature groups by their MI scores.
For each of the top-ranked feature groups, we additionally report whether its empirical usage is dominated by core or spurious annotations. Figure \ref{fig:top10_mi} shows the MI between feature groups and labels across all datasets, ranked by association strength. 
Across datasets, a substantial fraction of the feature groups most strongly associated with labels are spurious-dominant.
Among the top-20 feature groups ranked by MI, several exhibit higher spurious usage than core usage, demonstrating that strong statistical association with labels does not imply task relevance.
This finding highlights a key limitation of correlation-driven learning: features that are highly predictive may reflect dataset-specific shortcuts rather than causal or semantic signals.
Dataset-specific analyses further illustrate this phenomenon. 
In MM-IMDb, visually salient appearance and environment-related features exhibit higher MI with genre labels than several expression- or configuration-related features that are predominantly core. (Figure \ref{fig:label_association_mmimdb})
In CREMA-D, low-level visual attributes such as visual surface and visual effects show stronger label association than micro facial motions, despite the latter being more semantically aligned with emotional expression. (Figure \ref{fig:label_association_cremad})
In RAVDESS, the MI values of core-dominant expression features and spurious-dominant visual surface features are comparable, indicating that spurious cues can be as predictive as task-relevant ones even in controlled datasets. (Figure \ref{fig:label_association_ravdess})
Overall, this analysis demonstrates that mutual information alone is insufficient to distinguish core from spurious features.
The coexistence of high label association and spurious dominance motivates the need for explicit mechanisms that regulate internal feature reliance beyond correlation-based optimization.
\vspace{-1em}

\begin{figure}[!ht]
    \centering
    \includegraphics[width=\linewidth]{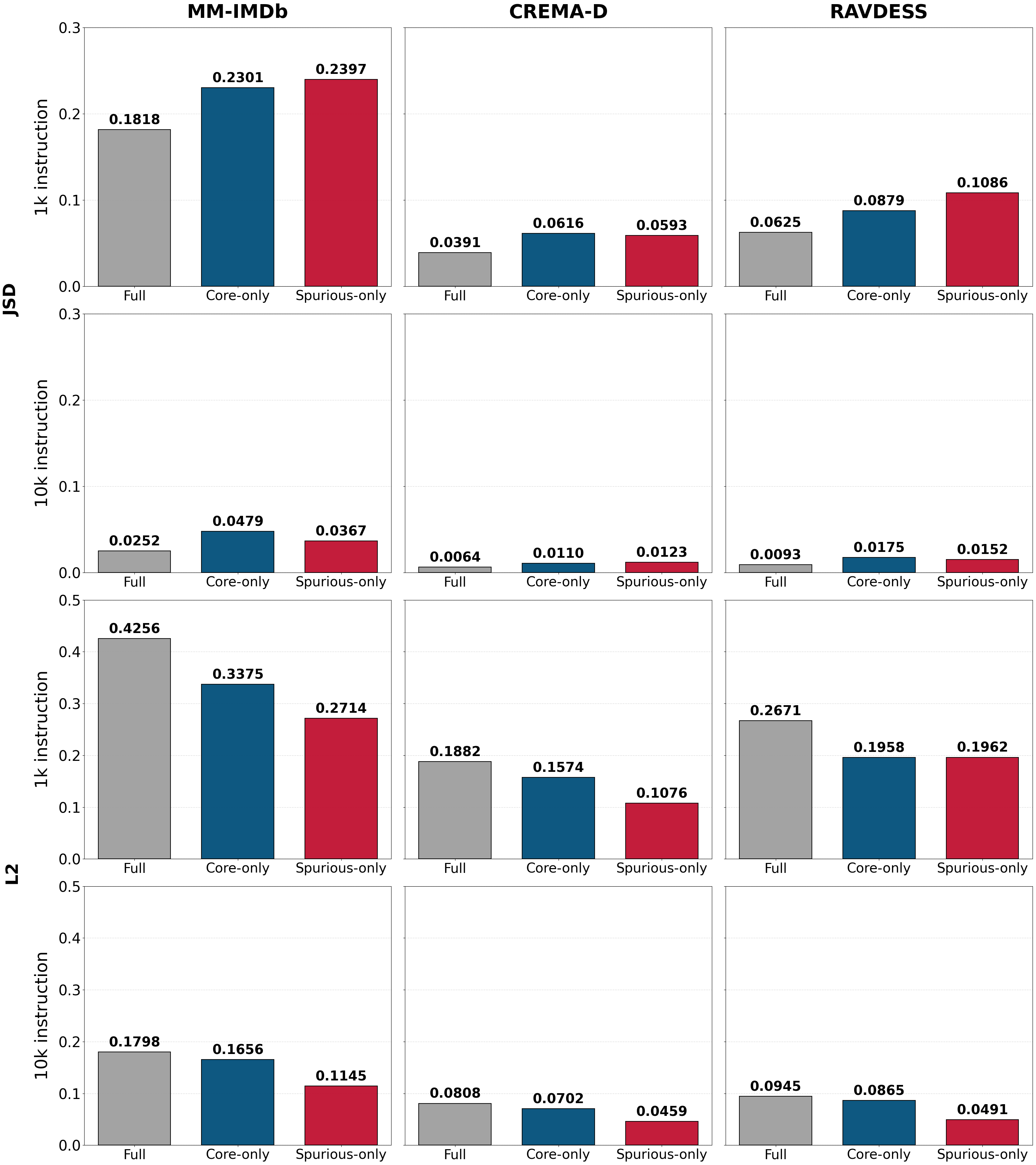}

    \caption{Label separability under different feature availability conditions. We measure Jensen-Shannon divergence (top) and $\ell_2$ distance (bottom) between label-wise average feature representations under three settings: \emph{Full} (all features), \emph{Core-only} (spurious features removed), and \emph{Spurious-only} (core features removed). Results are reported for 1k and 10k instruction variations across MM-IMDb, CREMA-D, and RAVDESS. Label separability differs only marginally between core-only and spurious-only conditions, indicating that spurious features alone can preserve substantial label structure.}
 
    \label{fig:label_separability}
\end{figure}

\FloatBarrier
\section{Label Separability Under Core and Spurious Feature Conditions}
\label{appendix:label_separability}
This appendix analyzes how well target labels are separable under different feature availability conditions, as shown in Figure \ref{fig:label_separability}.
Specifically, we compare label separability when using all available features (\emph{Full}), only core features (\emph{Core-only}), and only spurious features (\emph{Spurious-only}).
For each instruction, we construct a feature group occurrence vector by counting the feature groups activated under that instruction.
We then compute label-wise average feature vectors and measure inter-label separability using two complementary metrics:
Jensen-Shannon divergence (JSD), which captures distributional distinguishability, and $\ell_2$ distance, which reflects geometric separation in feature space.
We report results for 1k and 10k instruction variations to assess the effect of instruction diversity.
Across datasets and instruction scales, JSD-based separability differs only marginally between core-only and spurious-only conditions.
In several cases, particularly for MM-IMDb and RAVDESS under 1k instructions, spurious-only features yield equal or higher label separability than core-only features.
This indicates that spurious features alone can support non-trivial label discrimination.

As the number of instructions increases to 10k, overall JSD values decrease across all conditions, suggesting that increased instruction diversity dilutes label-specific feature concentration.
Notably, this degradation affects core-only and spurious-only conditions similarly, implying that neither feature subset uniquely preserves label separability at scale.
The $\ell_2$ distance analysis exhibits a consistent ordering across all datasets and instruction scales, with
\emph{Full} $>$ \emph{Core-only} $>$ \emph{Spurious-only}.
However, the magnitude of these differences is modest, and spurious-only representations remain comparably separable from a geometric perspective.

Together, these results show that label separability alone does not reliably distinguish core from spurious features.
Even when restricted to spurious features, the induced feature representations retain substantial label structure, explaining why correlation-driven learning can preferentially exploit spurious cues in practice.

\begin{figure}[t]
    \centering
    \subfigure[MM-IMDb (Qwen2.5-VL) - Stability]{
        \includegraphics[width=0.31\linewidth]{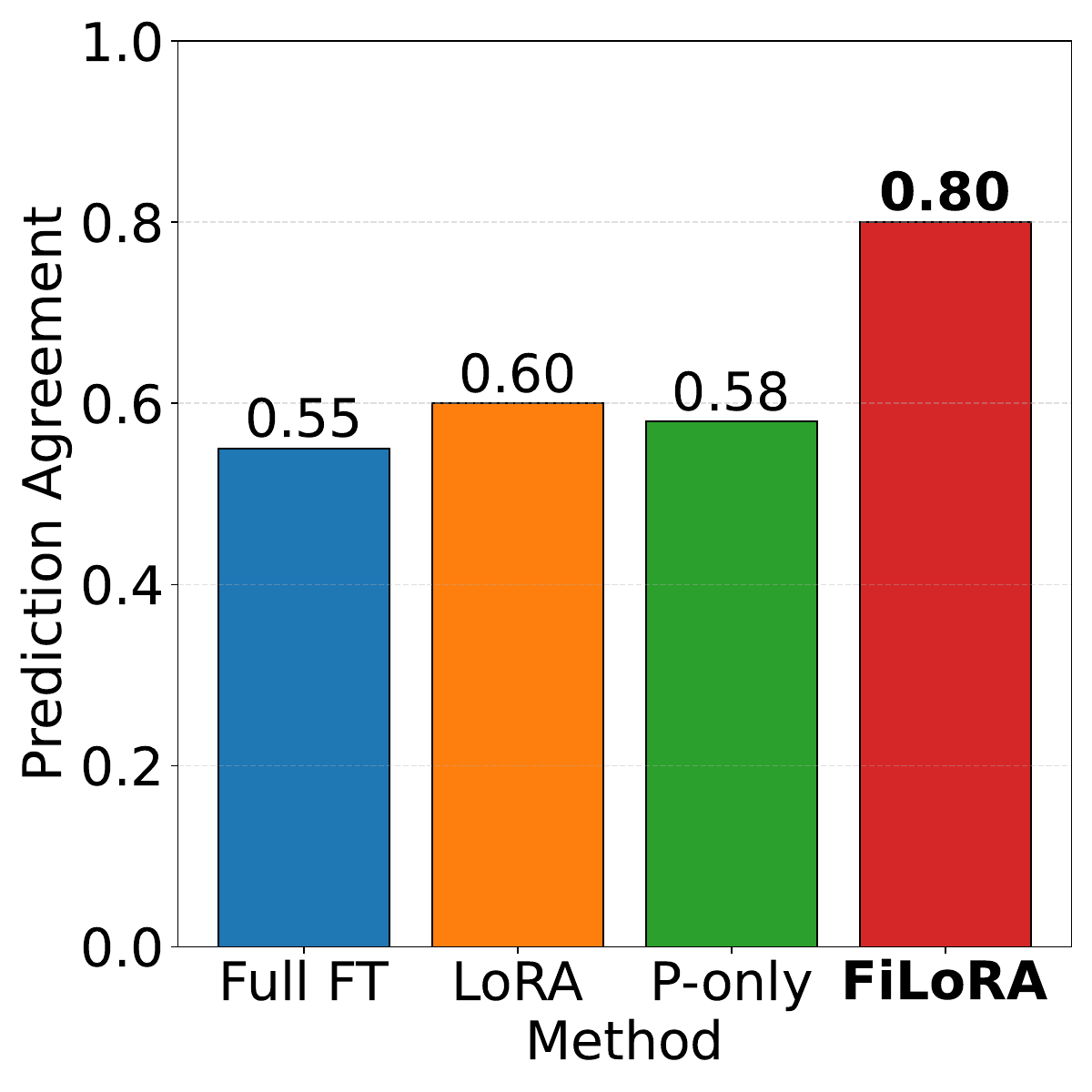}
        \label{fig:decision_stability_mmimdb}
    }
    \subfigure[CREMA-D (Qwen2.5-Omni) - Stability]{
        \includegraphics[width=0.31\linewidth]{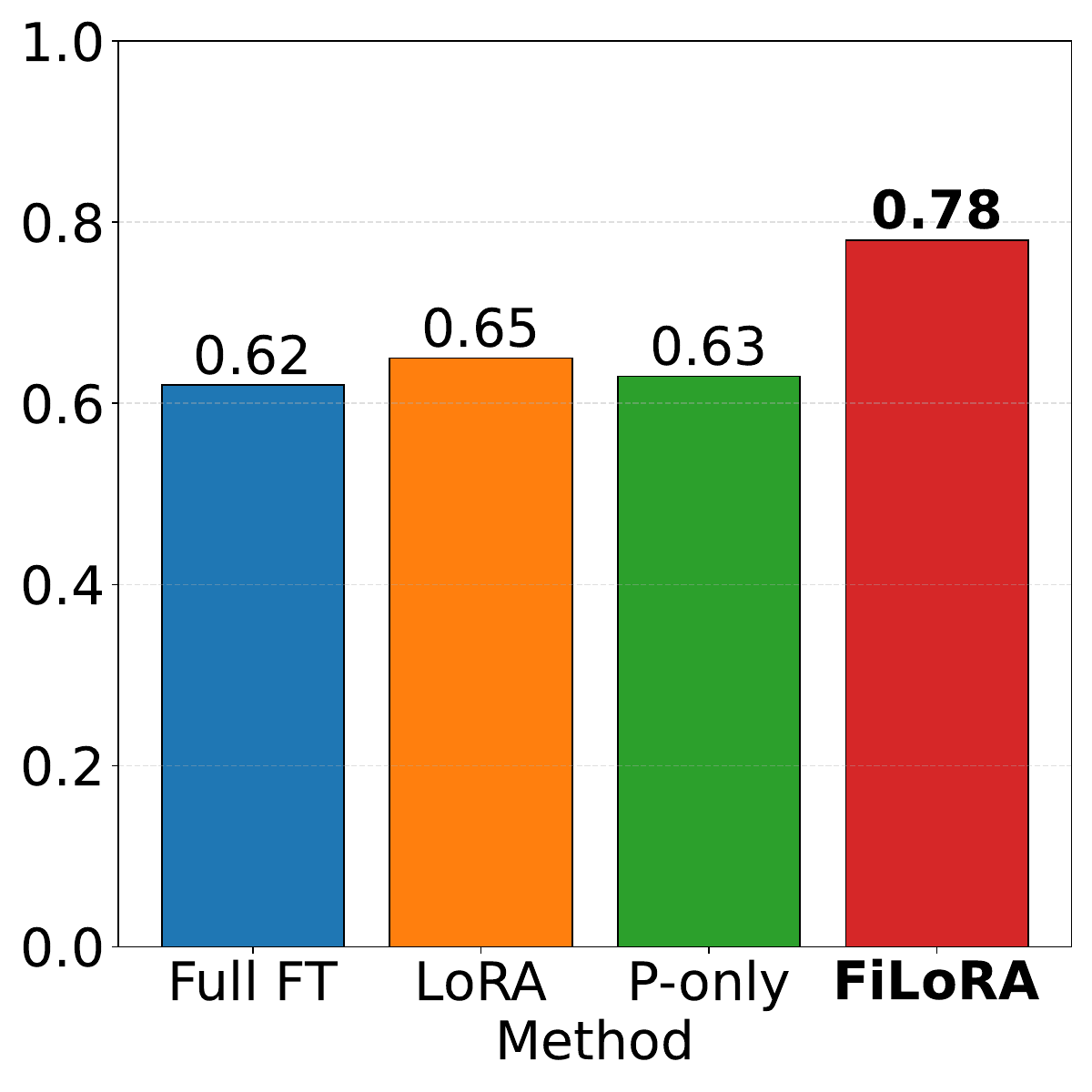}
        \label{fig:decision_stability_cremad}
    }
    \subfigure[RAVDESS (Qwen2.5-Omni) - Stability]{
        \includegraphics[width=0.31\linewidth]{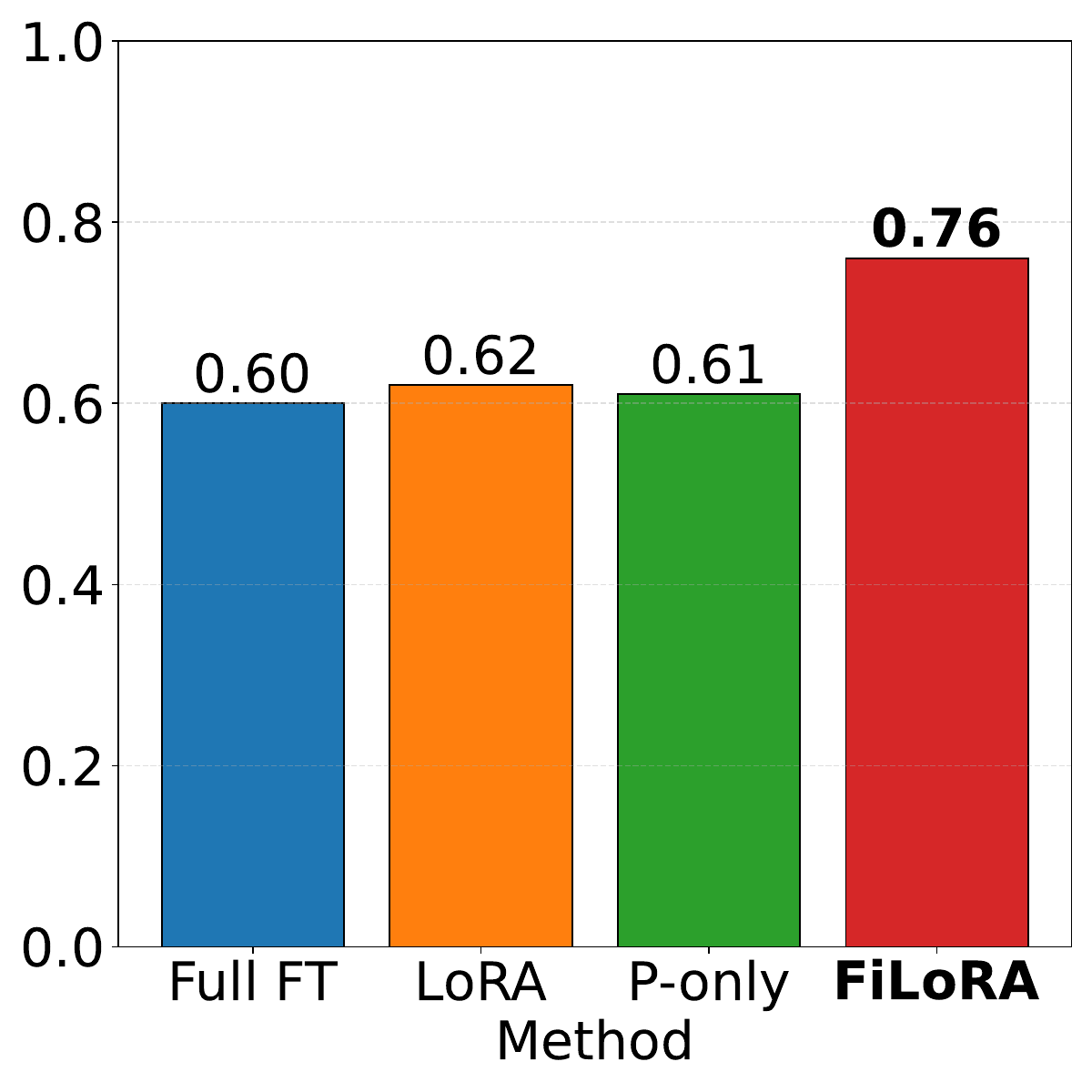}
        \label{fig:decision_stability_ravdess}
    }
    
    \subfigure[MM-IMDb (Qwen2.5-VL) - Degrad.]{
        \includegraphics[width=0.31\linewidth]{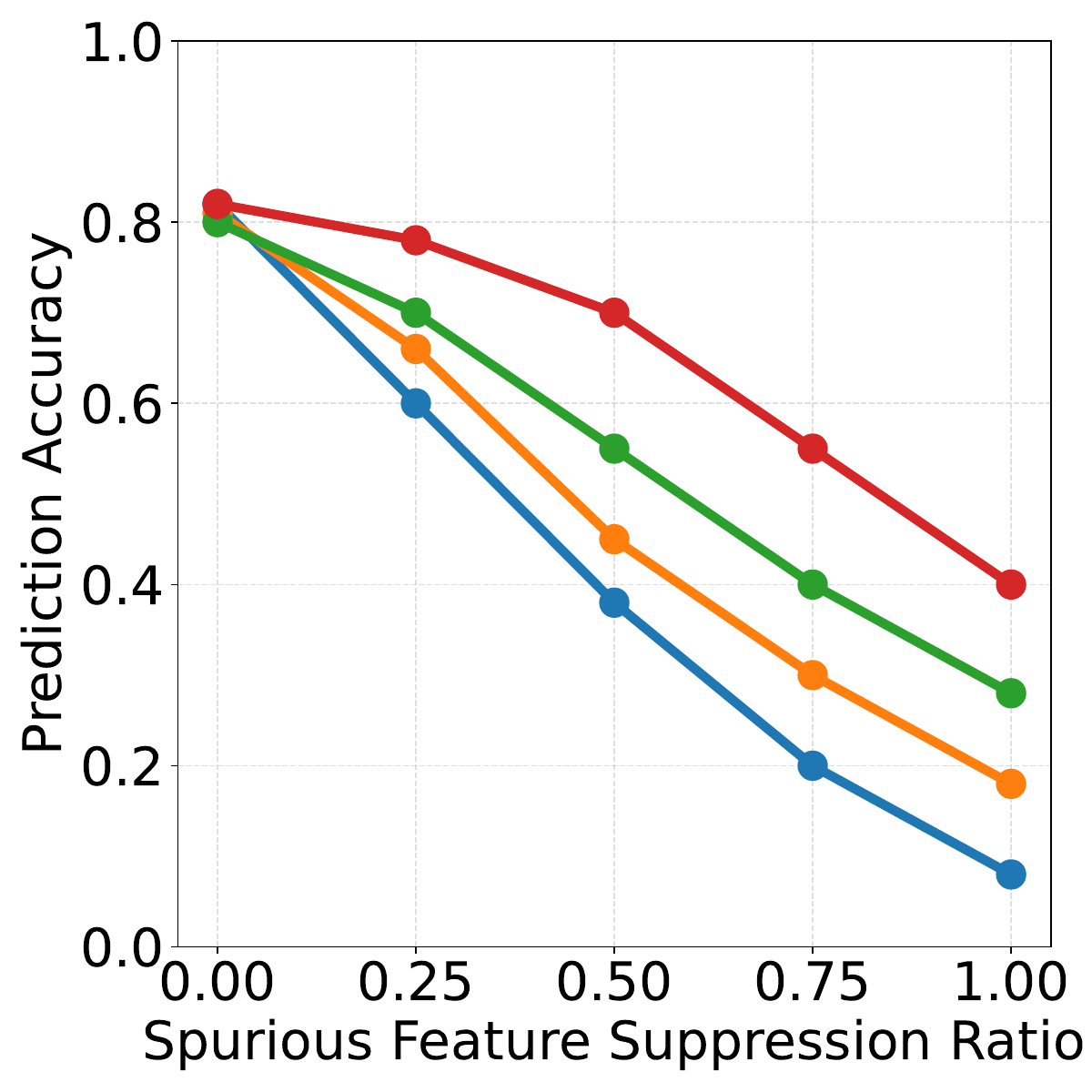}
        \label{fig:spurious_degradation_mmimdb}
    }
    \subfigure[CREMA-D (Qwen2.5-Omni) - Degrad.]{
        \includegraphics[width=0.31\linewidth]{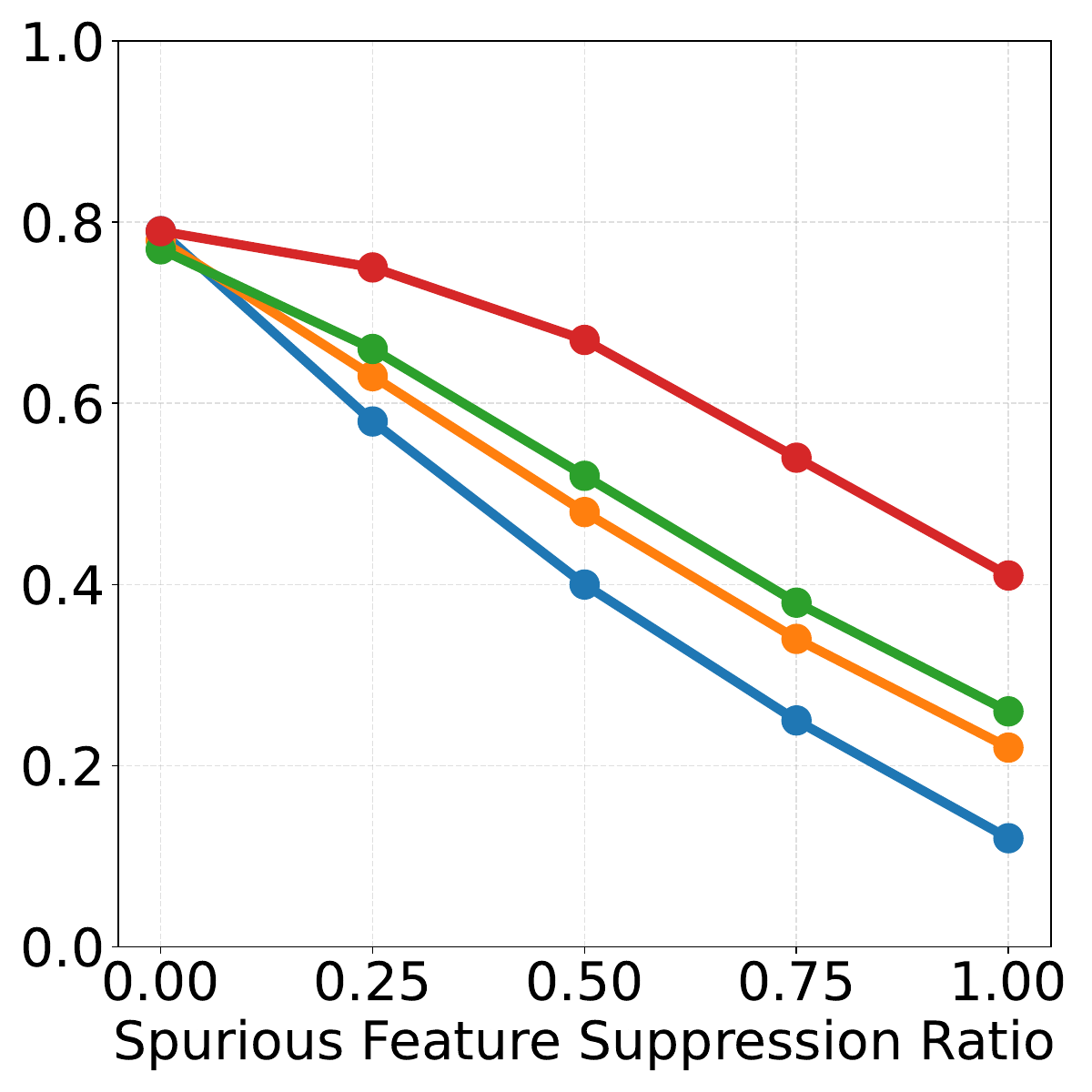}
        \label{fig:spurious_degradation_cremad}
    }
    \subfigure[RAVDESS (Qwen2.5-Omni) - Degrad.]{
        \includegraphics[width=0.31\linewidth]{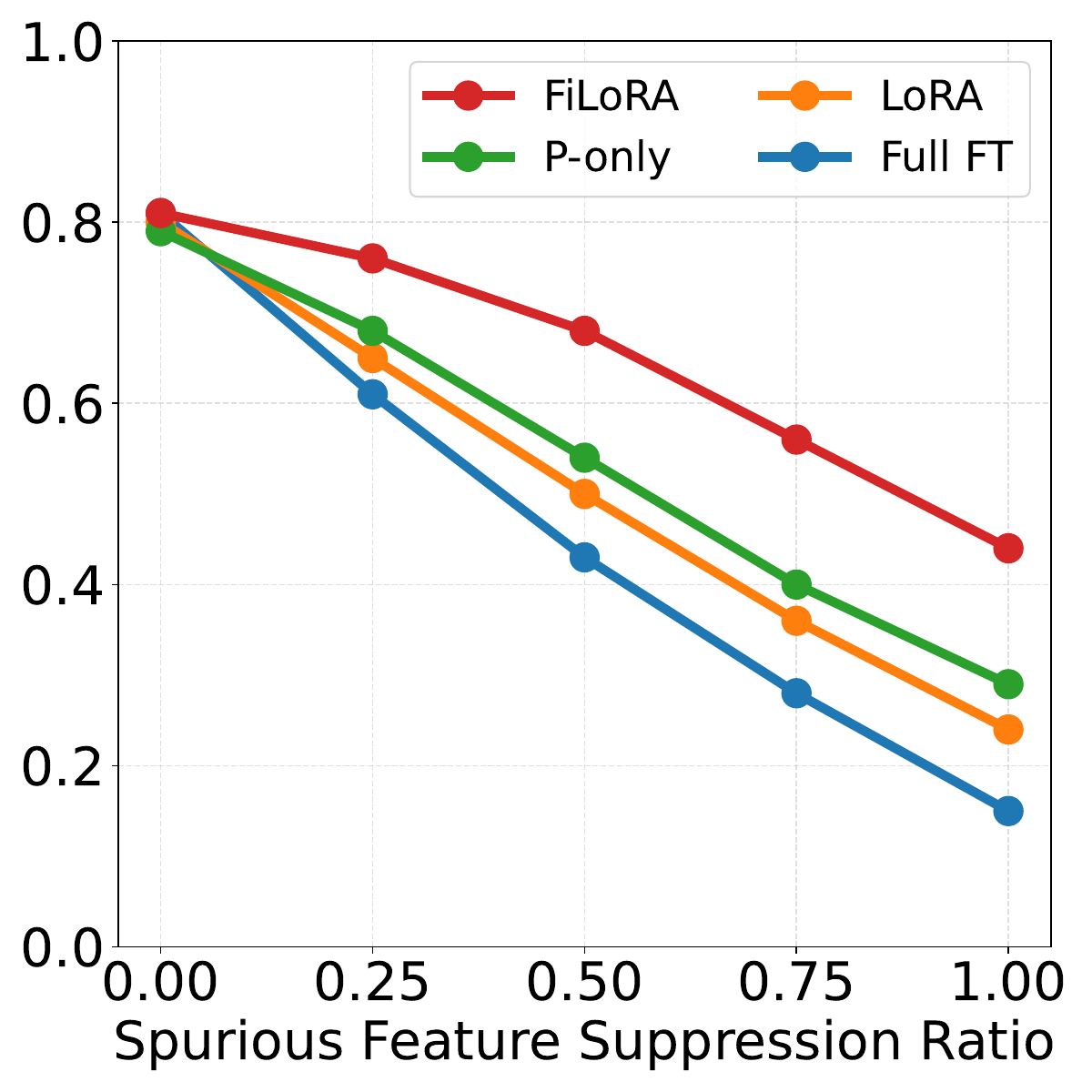}
        \label{spurious_degradation_ravdess}
    }
    
    \caption{
        Dataset-specific decision stability under spurious feature removal (top, abbreviated as Stability) and performance degradation under increasing spurious feature suppression (bottom, abbreviated as Degrad.). For decision stability, prediction agreement is measured before and after removing spurious features.
     All baselines are evaluated on dataset-appropriate backbones.
    Across datasets, FiLoRA consistently preserves higher stability than full fine-tuning, LoRA, and P-only (prompt-only) baselines. For performance degradation, accuracy is reported as a function of spurious feature suppression ratio. Although degradation rates differ across datasets, FiLoRA consistently exhibits more gradual performance decline than baseline methods.
    }
    \label{fig:robustness_dataset}
\end{figure}

\vspace{-0.9em}
\section{Dataset-Specific Robustness to Spurious Feature Interventions}
\label{appendix:robustness_dataset}
This appendix provides dataset-specific robustness analyses under spurious feature interventions, as shown in Figure \ref{fig:robustness_dataset}.
For each dataset, all baseline methods are evaluated on the appropriate backbone: Qwen2.5-VL for MM-IMDb and Qwen2.5-Omni for CREMA-D and RAVDESS.
We report decision stability under spurious feature removal and performance degradation under progressive spurious suppression.
While absolute robustness varies across datasets due to differences in spurious structure, all datasets exhibit consistent qualitative trends: FiLoRA preserves higher decision stability and degrades more gracefully than full fine-tuning, LoRA, and prompt-only baselines.

\vspace{-0.9em}
\section{Construction of Spurious Proxy Labels}
\label{appendix:proxy_labels}

To enable controlled analysis of feature reliance, we construct spurious proxy labels that approximate predictions made when the model relies primarily on a restricted subset of input features. These proxy labels are used exclusively as a training-time intervention and are never used as evaluation targets. For each dataset, we specify feature subsets that are known to be predictive yet semantically incidental to the task, and we intend the proxy labels to reflect reliance on these cues. 

In MM-IMDb, proxy labels are derived to emphasize visual appearance attributes of movie posters, such as color palettes, typography, and layout cues, while de-emphasizing semantic information conveyed by plot text.  In CREMA-D and RAVDESS, proxy labels are constructed to emphasize low-level acoustic or surface-level visual cues, such as pitch statistics, energy contours, or coarse visual attributes, while discouraging reliance on features directly encoding emotional semantics.

Proxy labels are generated offline using a frozen large vision--language model (LVLM), prompted to make predictions based only on the specified restricted cues. Because such models may imperfectly follow focus-ignore instructions, we treat the resulting proxy labels as spur-leaning supervision signals rather than pure spurious ground truth. To mitigate semantic leakage, we apply input-level restrictions appropriate to each modality and report diagnostic analyses that quantify proxy-core label alignment and potential leakage effects.

Both the original task labels and the proxy labels lie in the same label space, and no new prediction objectives are introduced. The proxy labels serve solely to induce controlled reliance on spurious cues during training, enabling us to study how instruction-conditioned gating modulates internal feature usage under fixed task semantics. At evaluation time, all models are assessed exclusively with respect to the original task labels.

\FloatBarrier
\section{FiLoRA Training Procedure}
\label{appendix:training_procedure}

This appendix provides a step-by-step description of the FiLoRA training procedure, complementing the methodological formulation presented in Section~\ref{sec:method}. 
Algorithm~\ref{alg:filora} summarizes how instruction-conditioned gating, grouped LoRA adaptation, and supervision routing are integrated during training.
Each training sample consists of multimodal input $x_i$, a natural language instruction $I_i$, the original task label $y_i$, a spurious proxy label $y_i^{(s)}$, and an experimental condition $c_i$.
The base multimodal model parameters remain frozen throughout training, and only grouped LoRA parameters and the instruction encoder are updated. At each training step, instructions are encoded into continuous control representations, which parameterize the instruction-conditioned gating function. These gates softly modulate the contribution of group-specific LoRA updates, enabling sample-wise and instruction-dependent control over internal computation paths.
The supervision signal is selected according to the experimental condition, but the condition itself is never provided to the model.
As a result, any learned shift in feature reliance must emerge from the interaction between instruction-conditioned gating and loss optimization, rather than from explicit conditioning on task identity.
A weak gate regularization term is optionally applied to stabilize instruction-conditioned reliance patterns and prevent degenerate gate behavior.
This regularization acts only as a soft inductive bias and does not impose hard constraints on gate activations.
All predictions are evaluated exclusively with respect to the original task labels, ensuring that the learning setup remains a single-task formulation.
Algorithm~\ref{alg:filora} is included to facilitate reproducibility and to clarify the end-to-end training flow of FiLoRA.

\begin{algorithm}[t]
\caption{FiLoRA Training with Instruction-Conditioned Gating}
\label{alg:filora}
\KwIn{
Training dataset $\{(x_i, I_i, y_i, y_i^{(s)}, c_i)\}_{i=1}^N$, \\
Pretrained multimodal model $f_\theta$ (frozen), \\
Grouped LoRA parameters $\{\Delta W_g\}_{g \in \mathcal{G}}$, \\
Instruction encoder $h_\phi$, \\
Gate regularization weight $\lambda$
}
\KwOut{Trained grouped LoRA parameters and instruction encoder}

\For{each training step}{
    Sample minibatch $\mathcal{B} \subset \{1,\dots,N\}$ \\
    \For{each sample $i \in \mathcal{B}$}{
    
        Encode instruction: $\mathbf{z}_i \leftarrow h_\phi(I_i)$ \\
        
        Compute gate vector: $\mathbf{g}_i \leftarrow \sigma(\mathbf{z}_i)$ \\
        
        Apply instruction-conditioned adaptation: \\
        \hspace{1em} $W' \leftarrow W + \sum_{g \in \mathcal{G}} g_{i,g}\,\Delta W_g$ \\
        
        Select supervision signal:
        \[
        y_i^{\text{target}} =
        \begin{cases}
        y_i, & \text{if } c_i \text{ emphasizes core features}, \\
        y_i^{(s)}, & \text{if } c_i \text{ emphasizes spurious features}
        \end{cases}
        \]
        
        Compute classification loss:
        \[
        \mathcal{L}_{\text{cls}}^{(i)} = \mathrm{CE}(f(x_i, I_i), y_i^{\text{target}})
        \]
        
        Compute gate regularization:
        \[
        \mathcal{L}_{\text{gate}}^{(i)} = \sum_{g \in \mathcal{G}} \alpha_g^{(i)}\, g_{i,g}
        \]
    }
    
    Update parameters by minimizing:
    \[
    \sum_{i \in \mathcal{B}} \left( \mathcal{L}_{\text{cls}}^{(i)} + \lambda \mathcal{L}_{\text{gate}}^{(i)} \right)
    \]
}

\end{algorithm}

 \begin{figure}[!h]
     \centering
         \subfigure[MM-IMDb]{
         \includegraphics[width=0.325\columnwidth]{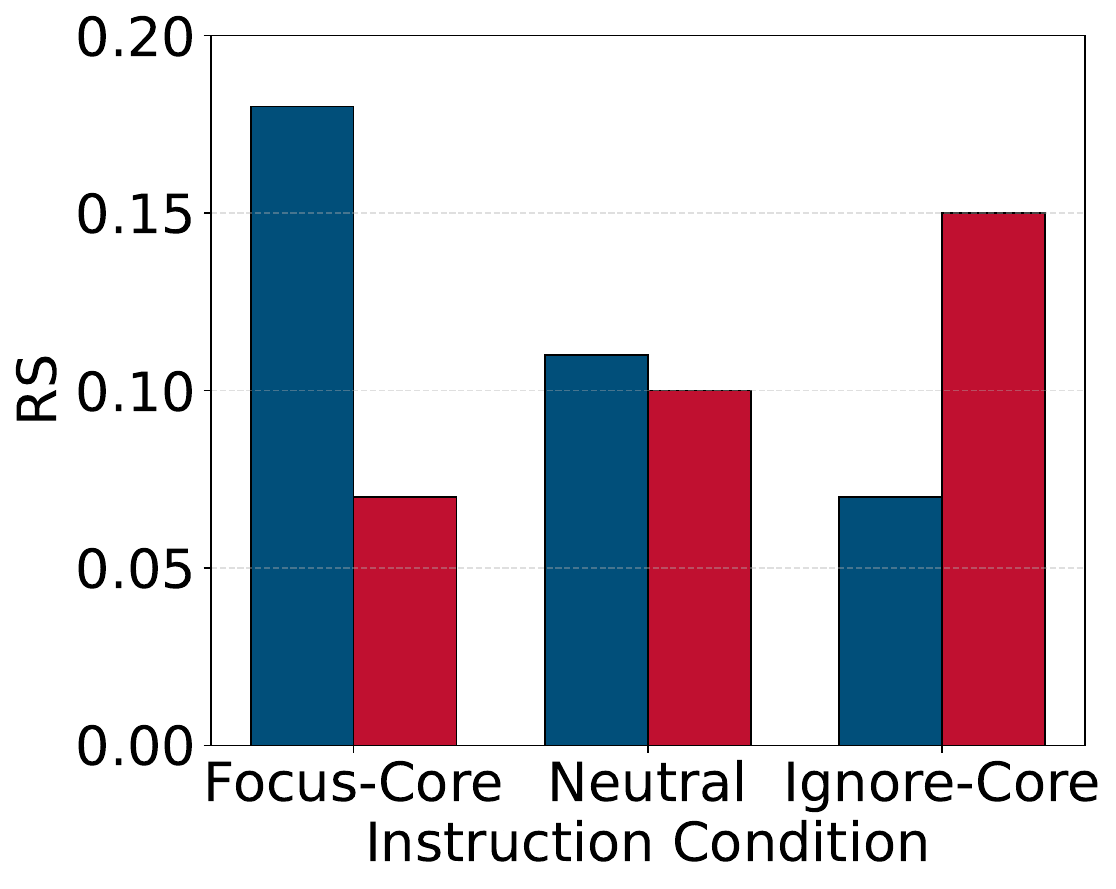}
     }
     \hspace{-1em}
     \subfigure[CREMA-D]{
         \includegraphics[width=0.325\columnwidth]{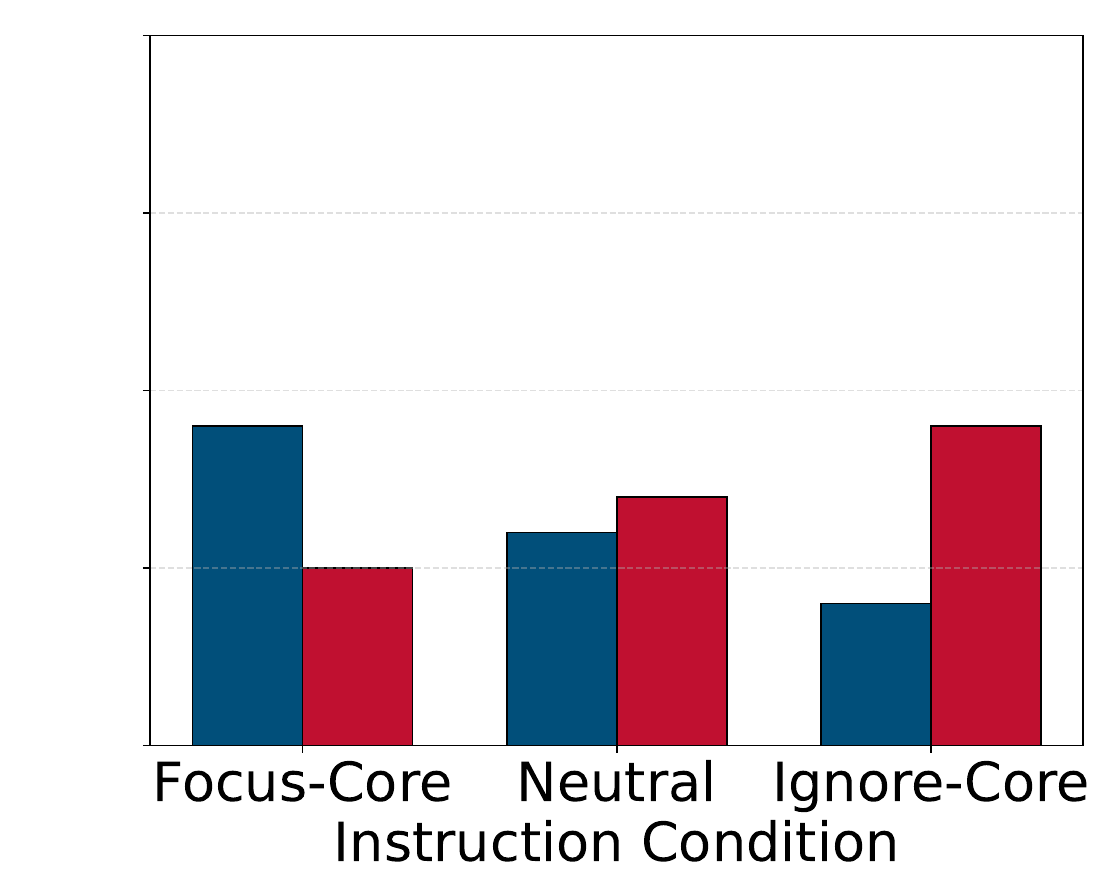}
     }
      \hspace{-1em}
     \subfigure[RAVDESS]{
         \includegraphics[width=0.325\columnwidth]{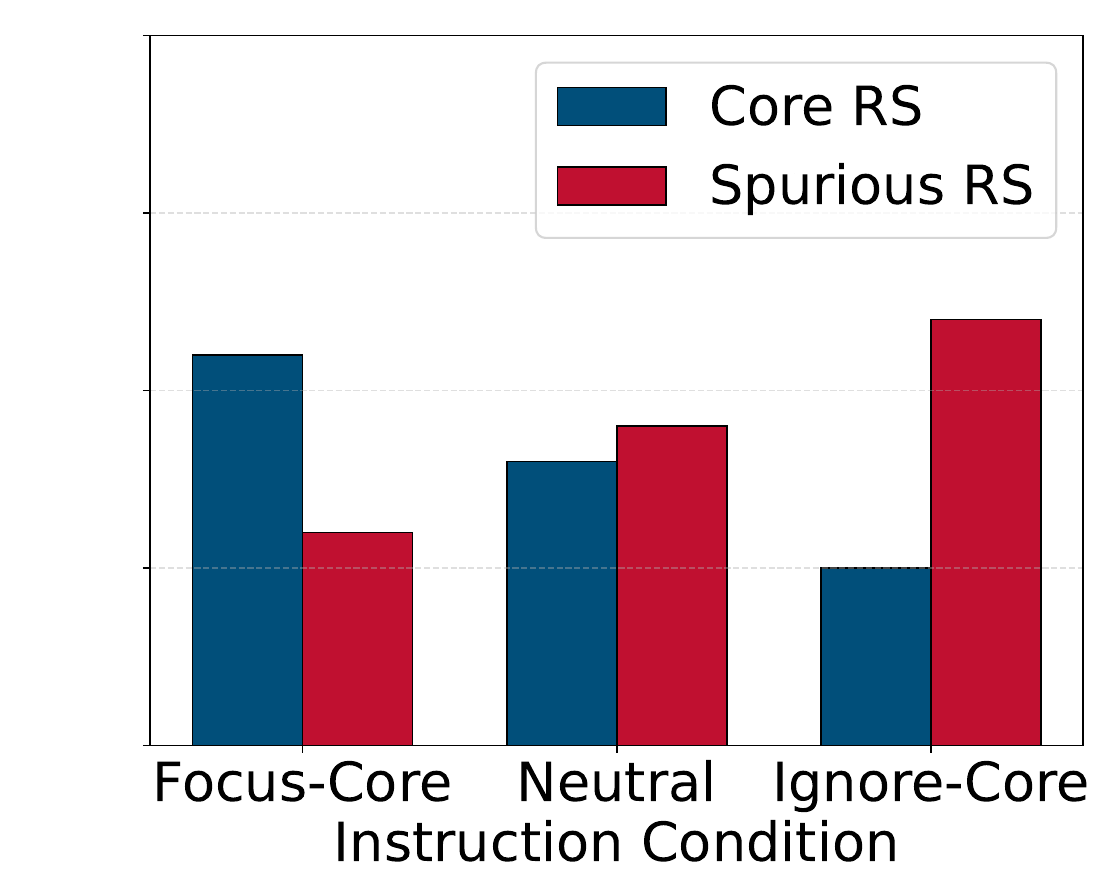}
     } 
     \vspace{-0.5em}
     \caption{Dataset-wise prediction sensitivity to instruction-conditioned gate perturbations across MM-IMDb, CREMA-D, and RAVDESS. Focus-core instructions increase sensitivity to core-related features, while ignore-core instructions shift sensitivity toward spurious features.}
     \label{fig:reliance_sensitivity_datasetwise}
 \end{figure}

\FloatBarrier
\section{Dataset-wise Reliance Sensitivity under Instruction-Conditioned Gating}
\label{appendix:rs_datasetwise} 
Figure~\ref{fig:reliance_sensitivity_datasetwise} shows that instruction-conditioned gating consistently redistributes prediction sensitivity between core and spurious features across datasets, providing direct evidence that reliance modulation produces functional changes in model behavior. The qualitative pattern mirrors the aggregate results in the main text, indicating that instruction-conditioned reliance modulation generalizes across modalities and dataset structures.

\section{Instruction Dataset used for FiLoRA}
\label{app:instruction_dataset}
Table \ref{tab:instruction_stat} shows the models used for generating instructions and the total number of instructions created for each dataset. We employed multimodal models to generate focus-ignore instructions by taking actual data samples as input. Specifically, we used Qwen2.5-VL-7B-Instruct for the vision-language dataset (MM-IMDb) and Qwen2.5-Omni-7B for the audio--visual datasets (CREMA-D and RAVDESS). When training FiLoRA, 1,000 instructions were used for each dataset.

\begin{table}[h]
\caption{The number of instructions for each dataset. We used the Qwen2.5-VL for MM-IMDb and Qwen2.5-Omni for CREMA-D and RAVDESS. 
Instructions were generated by multimodal models using actual data samples as input to create focus-ignore instructions.}
\centering
\resizebox{0.5\columnwidth}{!}{%
\begin{tabular}{llc}
\toprule
\textbf{Dataset} & \textbf{Model} & \textbf{Total number of instructions} \\
\midrule
MM-IMDb  & Qwen2.5-VL-7B-Instruct   & 1,000 \\
CREMA-D  & Qwen2.5-Omni-7B &  1,000 \\
RAVDESS  & Qwen2.5-Omni-7B &  1,000 \\
\bottomrule
\end{tabular}%
}
\label{tab:instruction_stat}
\end{table}

\section{Dataset Selection Criteria}
\label{appendix:dataset_criteria}
We select datasets according to three principled criteria that are necessary to support causal analysis of instruction-conditioned controllability. 
First, datasets must exhibit identifiable deterministic feature groups, such that low-level observable cues(e.g., acoustic traits and visual attributes) can be reliably mapped to higher-level modality or feature groups. This property ensures causally tracing the instruction-reliance pathway and also improves experimental reproducibility.
Second, datasets should support diverse feature groups, enabling systematic and programmatic expansion of instruction templates. This allows us to generate large families of semantically equivalent instructions while reducing sensitivity to specific wording choices. Unlike feature removal or corruption-based analysis, FiLoRA enables continuous, instruction-conditioned modulation of internal reliance without altering the input distribution.
Third, we require tasks with a fixed label space and stable objective, ensuring that variations across instruction conditions reflect changes in internal reliance rather than shifts in task semantics. 
In summary, datasets are chosen for (1) deterministic feature groups enabling causal traceability and reproducibility, and (2) compatibility with programmatic instruction expansion. We therefore focus on datasets where feature groups can be reasonably distinguished, as this supports clearer analysis and interpretation of instruction-conditioned changes in internal reliance.

\section{Implementation Details}
\label{app:implementation_details}
All experiments were conducted on a single \textbf{NVIDIA H100 NVL GPU (94GB memory)} using mixed-precision training (bfloat16).
We used the AdamW optimizer with a batch size of 16, learning rates of $2 \times 10^{-4}$ for grouped LoRA parameters and
$1 \times 10^{-4}$ for the instruction encoder, and a weight decay of $1 \times 10^{-2}$.
Gradient norms were clipped at 1.0.
Backbone multimodal models were kept frozen throughout training.

FiLoRA employs grouped LoRA adaptation with rank 8 and scaling factor 16.
Instruction-conditioned gating produces a per-instruction gate vector with dimension equal to the number of LoRA groups.
The instruction encoder is implemented as a lightweight 2-layer Transformer with hidden dimension 512 and GELU activations.
A weak gate regularization term is applied to prevent degenerate gating behavior.

Training required approximately 2--3 hours per run depending on the dataset and backbone, resulting in a total compute budget
of roughly \textbf{120 GPU-hours} on H100 NVL GPUs.
All random seeds were fixed to ensure reproducibility.

\section{Extended Generative Evaluation on Flickr30K}
\label{appendix:caption_taxonomy_examples}

We provide additional qualitative and quantitative results for instruction-conditioned caption generation on Flickr30K using extended instruction taxonomies beyond the main paper settings.

While the main paper focuses on Semantic Focus, Appearance Focus, and Appearance Suppress conditions, we additionally evaluate Conceptual Focus, Behavior Focus, Background Suppress, and Demographic Suppress instructions to analyze whether FiLoRA generalizes to broader forms of feature-level control during autoregressive generation.

\begin{table}[t]
\caption{
Extended Flickr30K generative evaluation across diverse instruction conditions.
FiLoRA maintains competitive caption quality while enabling controllable shifts in feature reliance across semantic, behavioral, contextual, and demographic dimensions.
}
\centering
\small
\setlength{\tabcolsep}{4.5pt}
\renewcommand{\arraystretch}{0.95}

\begin{tabular}{lccc}
\toprule
\textbf{Instruction Condition} & \textbf{CIDEr} & \textbf{CLIPScore} & \textbf{GMR} \\
\midrule
Semantic Focus       & 112.4 & 0.823 & 0.41 \\
Appearance Focus     & 109.7 & 0.817 & 0.39 \\
Appearance Suppress  & 113.1 & 0.826 & 0.43 \\
Conceptual Focus     & 111.8 & 0.821 & 0.40 \\
Behavior Focus       & 110.9 & 0.819 & 0.38 \\
Background Suppress  & 113.5 & 0.828 & 0.44 \\
Demographic Suppress & 112.7 & 0.824 & 0.42 \\
\bottomrule
\end{tabular}

\label{tab:flickr_captioning_full}
\end{table}

\begin{figure}[!t]
    \centering
    \includegraphics[width=\columnwidth]{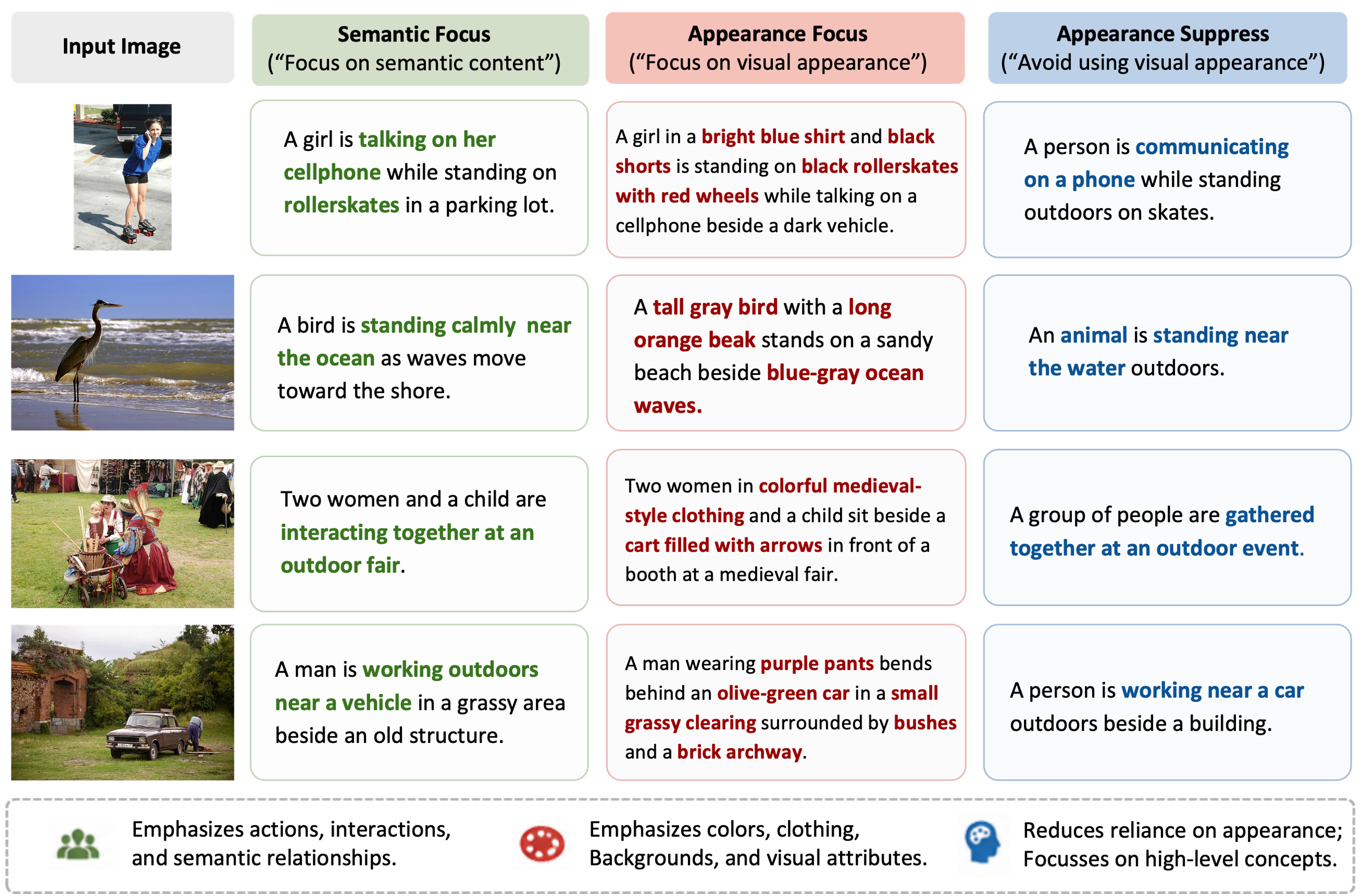}
    \vspace{-1.5em}
    \caption{Additional qualitative examples for Semantic Focus, Appearance Focus, and Appearance Suppress instructions on Flickr30K.
    FiLoRA consistently modifies caption emphasis according to instruction semantics while preserving coherent scene understanding.}
    \label{fig:flickr_caption_examples_append}
\end{figure}

Table~\ref{tab:flickr_captioning_full} shows that FiLoRA preserves stable caption quality across diverse instruction conditions while maintaining substantial gate modulation behavior.
Importantly, controllable reliance shifts extend beyond appearance-oriented interventions to higher-level semantic, behavioral, contextual, and demographic control settings.

Figure~\ref{fig:flickr_caption_examples_append} presents additional qualitative examples under Semantic Focus, Appearance Focus, and Appearance Suppress instructions.
The generated captions exhibit systematic shifts in semantic emphasis: semantic-focused instructions prioritize interactions and scene meaning, appearance-focused instructions increase reliance on visual attributes and descriptive details, while appearance-suppress instructions reduce dependence on low-level appearance cues and produce more abstract scene descriptions.

\begin{figure}[h]
    \centering
    \includegraphics[width=\columnwidth]{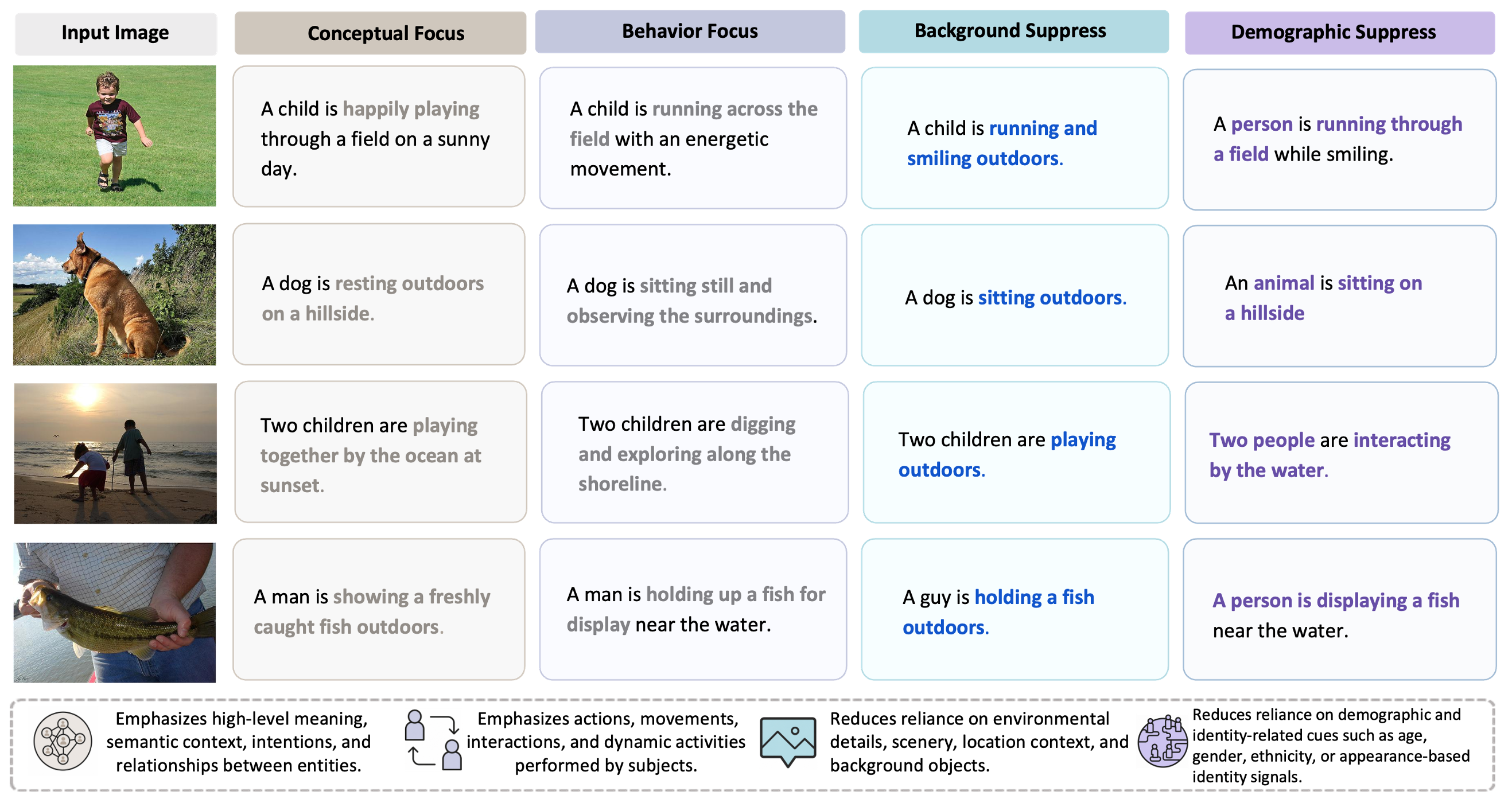}
    \vspace{-1.5em}
    \caption{Additional qualitative examples for extended instruction taxonomies on Flickr30K, including Conceptual Focus, Behavior Focus, Background Suppress, and Demographic Suppress conditions.
    FiLoRA generalizes beyond fixed focus/suppress templates and supports diverse forms of semantic and contextual reliance control during caption generation.}
    \label{fig:flickr_caption_examples_append_4ver}
\end{figure}

Figure~\ref{fig:flickr_caption_examples_append_4ver} further extends this analysis to Conceptual Focus, Behavior Focus, Background Suppress, and Demographic Suppress instructions.
These examples demonstrate that FiLoRA supports broader forms of controllable reliance modulation beyond simple focus/suppress formulations.
Conceptual-focused instructions emphasize high-level scene meaning and abstract interpretation, behavior-focused instructions prioritize actions and interactions, background-suppress instructions reduce contextual scene dependence, and demographic-suppress instructions produce more identity-agnostic descriptions while preserving overall semantic coherence.

Overall, these results suggest that instruction-conditioned reliance modulation extends beyond controlled classification settings and supports structured control over feature-level reliance in open-ended multimodal generation under diverse semantic instruction conditions.

\section{Sensitivity to Proxy Supervision Quality}
\label{appendix:proxy_analysis}
We next analyze the relationship between proxy supervision and original task labels to evaluate whether FiLoRA relies on trivial relabeling effects or genuinely distinct intervention signals.

\begin{table}[t]
\caption{
Agreement between proxy supervision labels and original task labels across datasets. 
Agreement is measured as label-level correspondence between proxy supervision and original task labels. 
Moderate agreement indicates that proxy labels remain correlated with task-relevant information while still differing from the original supervision targets.
}
\centering
\small
\setlength{\tabcolsep}{6pt}
\renewcommand{\arraystretch}{0.95}

\begin{tabular}{lc}
\toprule
\textbf{Dataset} & \textbf{Proxy / Task Agreement} \\
\midrule
MM-IMDb & 0.68 \\
CREMA-D & 0.61 \\
RAVDESS & 0.59 \\
\bottomrule
\end{tabular}

\label{tab:proxy_agreement}
\end{table}
Table~\ref{tab:proxy_agreement} shows that proxy supervision exhibits moderate agreement with the original task labels across datasets. This behavior is expected for shortcut-associated cues: proxy labels remain partially correlated with task-relevant information while still capturing distinct sources of evidence. Importantly, agreement remains substantially below full correspondence, indicating that proxy supervision does not simply reproduce the original targets. Instead, proxy labels act as distinct intervention signals that bias the model toward alternative feature pathways during training. All evaluations are performed using the original task labels rather than the proxy supervision targets. As a result, the observed changes in controllability, robustness, and reliance sensitivity reflect shifts in feature reliance behavior rather than alignment to proxy labels themselves. These results suggest that FiLoRA remains effective under imperfect and partially correlated supervision signals, supporting proxy supervision as a practical mechanism for controllable reliance intervention under fixed task semantics.

\vspace{-0.5em}

\section{Generalization Across Architectures and Scales}
\label{appendix:architecture_generalization}

We evaluate whether FiLoRA exhibits consistent reliance modulation across multimodal backbones with different architectures, scales, and inductive biases.
For vision--language experiments on MM-IMDb, we evaluate Qwen2.5-VL-7B, Qwen2.5-VL-32B, and Gemma-3n-E4B.
For audio--visual experiments on CREMA-D and RAVDESS, we evaluate Qwen2.5-Omni-7B, AV-HuBERT, and AudioCLIP.
Table~\ref{tab:av_generalization} reports backbone-specific Core RS and Spurious RS under neutral instructions, together with GMR measured using contrasting focus and ignore instructions.
Although absolute RS values vary across architectures and model scales, the relative balance between core- and spurious-associated sensitivity remains qualitatively consistent across backbones within each dataset.
Specifically, MM-IMDb exhibits greater sensitivity to spurious-associated groups, CREMA-D shows stronger core-associated sensitivity, and RAVDESS exhibits a comparatively balanced reliance profile.
These findings indicate that dataset-specific reliance patterns are preserved across heterogeneous backbones, while GMR demonstrates gate responsiveness to contrasting instructions.
Together with the instruction-conditioned analyses in Section~5.2, these results support the applicability of FiLoRA across different multimodal architectures and scales.

\begin{table}[t]
\caption{
Generalization of FiLoRA across heterogeneous multimodal architectures and model scales.
Core RS and Spurious RS are evaluated under neutral instructions to characterize backbone-specific reliance profiles, while GMR measures gate modulation between contrasting focus and ignore instructions.
}

\centering
\small
\setlength{\tabcolsep}{4pt}
\renewcommand{\arraystretch}{0.95}

\begin{tabular}{llccc}
\toprule
\textbf{Dataset} & \textbf{Backbone} & \textbf{Core RS} & \textbf{Spurious RS} & \textbf{GMR} \\
\midrule
MM-IMDb & Qwen2.5-VL-7B  & 0.11 & 0.16 & 0.43 \\
MM-IMDb & Qwen2.5-VL-32B & 0.17 & 0.21 & 0.51 \\
MM-IMDb & Gemma-3n-E4B   & 0.10 & 0.14 & 0.39 \\
\midrule
CREMA-D & Qwen2.5-Omni-7B & 0.14 & 0.11 & 0.45 \\
CREMA-D & AV-HuBERT       & 0.15 & 0.10 & 0.47 \\
CREMA-D & AudioCLIP       & 0.14 & 0.11 & 0.46 \\
\midrule
RAVDESS & Qwen2.5-Omni-7B & 0.13 & 0.12 & 0.44 \\
RAVDESS & AV-HuBERT       & 0.13 & 0.11 & 0.43 \\
RAVDESS & AudioCLIP       & 0.12 & 0.11 & 0.42 \\
\bottomrule
\end{tabular}
\label{tab:av_generalization}
\end{table}

\section{Dataset-Wise Reliance Profiles} 
\label{appendix:dataset_reliance_profiles}

\begin{figure}[t]
    \centering
    \includegraphics[width=\columnwidth]{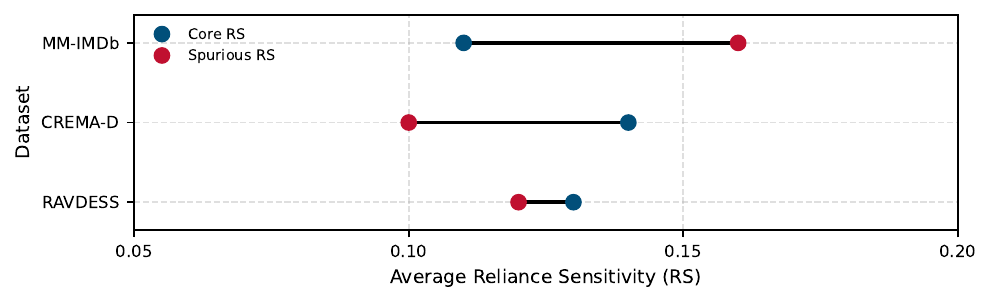}
    \vspace{-2em}
    \caption{
Dataset-wise reliance profiles measured by average RS. Each point reports prediction sensitivity to gate perturbations for core and spurious feature groups under neutral instructions.
    }
    \label{fig:dataset_reliance_profiles}
\end{figure}

\begin{table}[!h]
\caption{Dataset-wise reliance profiles under neutral instructions using the reference backbone for
each dataset. Values represent average prediction sensitivity to core- and spurious-associated gates.
C/S denotes the ratio of Core RS to Spurious RS.}
\centering
\resizebox{0.5\columnwidth}{!}{%
\begin{tabular}{lccc}
\toprule
\textbf{Dataset} & \textbf{Core RS} & \textbf{Spurious RS} & \textbf{Core / Spurious  ratio} \\
\midrule
MM-IMDb  & 0.11 & 0.16 & 0.69 \\
CREMA-D  & 0.14 & 0.11 & 1.27 \\
RAVDESS  & 0.13 & 0.12 & 1.08 \\
\bottomrule
\end{tabular}%
}
\label{tab:dataset_reliance}
\end{table}

We next analyze how instruction-conditioned reliance modulation manifests across datasets with different intrinsic feature compositions.
Figure~\ref{fig:dataset_reliance_profiles} reports average RS for core and spurious feature groups under neutral instructions. Numerical summaries of these dataset-wise reliance sensitivities are provided in Table~\ref{tab:dataset_reliance} in the Appendix.
Different datasets exhibit clearly distinct reliance structures. 
For MM-IMDb, spurious-related feature groups exhibit comparatively higher sensitivity, reflecting the strong association between visual appearance cues and genre labels in movie posters. 
In contrast, CREMA-D exhibits substantially higher sensitivity to core-related pathways, consistent with emotion recognition relying more heavily on semantic facial and acoustic signals. 
RAVDESS lies between these extremes, showing a comparatively balanced reliance structure.
These results suggest that FiLoRA does not enforce a fixed or uniform reliance bias across tasks. 
Instead, it exposes and modulates dataset-specific reliance patterns, enabling controlled analysis of when and where shortcut-associated features dominate model behavior.
Additional analyses of feature-label mutual information, shortcut-associated feature concentration, and label separability are provided in Appendix~\ref{appendix:label_feature_mi} and Appendix~\ref{appendix:label_separability}.

\section{Broader Impact}
This work studies how multimodal models rely on different internal feature pathways under natural language instructions and introduces FiLoRA as a mechanism for controllable reliance modulation. 
By enabling models to reduce dependence on shortcut-associated features while preserving task semantics, FiLoRA may contribute to more robust, interpretable, and controllable multimodal systems in domains such as multimedia understanding, affect recognition, and human-centered AI applications.
More broadly, this work argues that multimodal instruction following should involve not only controlling outputs, but also understanding and regulating how models internally use information. 
We hope this perspective encourages future research on controllable internal mechanisms for large multimodal models, including methods for feature-level interpretability, reliance auditing, and intervention-based robustness analysis.
At the same time, mechanisms that enable fine-grained control over internal feature usage may introduce risks if misused. 
In particular, instruction-conditioned reliance steering could potentially be exploited to amplify harmful biases, suppress relevant evidence, or manipulate downstream decision-making behavior in sensitive applications.
We partially mitigate these concerns through several design choices. 
FiLoRA does not modify model inputs or task objectives, and all evaluations are conducted using the original ground-truth labels. 
Proxy supervision is used only as a training-time intervention signal and is never exposed during inference. 
Furthermore, our work focuses on analyzing and regulating feature reliance rather than optimizing solely for task performance improvements.
Overall, we view FiLoRA as a step toward more accountable multimodal systems in which instruction following is coupled with explicit understanding of how internal feature reliance is formed and controlled.

\section{Limitations}
FiLoRA currently relies on semantically motivated grouping of LoRA updates to enable interpretable control over feature pathways. 
Although our experiments show that semantically aligned grouping consistently improves controllability compared to coarse or random partitioning, determining optimal grouping structures for large-scale multimodal systems remains an open design question. 
Future work may explore automated or dynamically learned grouping strategies across architectures and tasks.

In addition, our experiments focus primarily on controlled multimodal settings where feature groups can be more clearly identified and analyzed. 
While this setup enables precise evaluation of instruction-conditioned reliance modulation, extending such control to highly entangled real-world environments remains an important direction for future work.

Finally, although FiLoRA generalizes across both discriminative and generative multimodal tasks, we have not yet evaluated its behavior in more complex multimodal reasoning, dialogue, or agentic systems. 
Understanding how reliance modulation interacts with long-horizon reasoning and iterative decision making remains an important area for future research.


\newpage
\section*{NeurIPS Paper Checklist}

\begin{enumerate}

\item {\bf Claims}
    \item[] Question: Do the main claims made in the abstract and introduction accurately reflect the paper's contributions and scope?
    \item[] Answer: \answerYes{} 
    \item[] Justification: The abstract and introduction accurately summarize the paper’s main contributions, and the corresponding claims are supported by the experiments and analyses in Sections 3–5.

\item {\bf Limitations}
    \item[] Question: Does the paper discuss the limitations of the work performed by the authors?
    \item[] Answer: \answerYes{} 
    \item[] Justification: The paper discusses the limitations of the proposed framework in the limitation section.

\item {\bf Theory assumptions and proofs}
    \item[] Question: For each theoretical result, does the paper provide the full set of assumptions and a complete (and correct) proof?
    \item[] Answer: \answerNA{} 
    \item[] Justification: The paper does not present formal theoretical theorems or proofs, but provides mathematical formulations and empirical analyses for the proposed framework.

    \item {\bf Experimental result reproducibility}
    \item[] Question: Does the paper fully disclose all the information needed to reproduce the main experimental results of the paper to the extent that it affects the main claims and/or conclusions of the paper (regardless of whether the code and data are provided or not)?
    \item[] Answer: \answerYes{} 
    \item[] Justification: The paper provides sufficient details to reproduce the main experiments, including datasets, models, training setup, and evaluation metrics.

\item {\bf Open access to data and code}
    \item[] Question: Does the paper provide open access to the data and code, with sufficient instructions to faithfully reproduce the main experimental results, as described in supplemental material?
    \item[] Answer: \answerYes{} 
    \item[] Justification: Open access to the code and reproduction details are provided in the supplementary material.

\item {\bf Experimental setting/details}
    \item[] Question: Does the paper specify all the training and test details (e.g., data splits, hyperparameters, how they were chosen, type of optimizer) necessary to understand the results?
    \item[] Answer: \answerYes{} 
    \item[] Justification: The paper specifies the datasets, training setup, model configurations, evaluation protocols, and implementation details necessary to understand the experimental results.

\item {\bf Experiment statistical significance}
    \item[] Question: Does the paper report error bars suitably and correctly defined or other appropriate information about the statistical significance of the experiments?
    \item[] Answer: \answerYes{} 
    \item[] Justification: The paper focuses on consistent empirical trends across datasets and model backbones.

\item {\bf Experiments compute resources}
    \item[] Question: For each experiment, does the paper provide sufficient information on the computer resources (type of compute workers, memory, time of execution) needed to reproduce the experiments?
    \item[] Answer: \answerYes{} 
    \item[] Justification: The paper provides implementation and compute environment details used for training and evaluation, primarily in the appendix.
    
\item {\bf Code of ethics}
    \item[] Question: Does the research conducted in the paper conform, in every respect, with the NeurIPS Code of Ethics \url{https://neurips.cc/public/EthicsGuidelines}?
    \item[] Answer: \answerYes{} 
    \item[] Justification: The research complies with the NeurIPS Code of Ethics.

\item {\bf Broader impacts}
    \item[] Question: Does the paper discuss both potential positive societal impacts and negative societal impacts of the work performed?
    \item[] Answer: \answerYes{} 
    \item[] Justification: The paper discusses both the potential benefits and risks of controllable feature reliance in multimodal models.
    
\item {\bf Safeguards}
    \item[] Question: Does the paper describe safeguards that have been put in place for responsible release of data or models that have a high risk for misuse (e.g., pre-trained language models, image generators, or scraped datasets)?
    \item[] Answer: \answerNA{} 
    \item[] Justification: This paper poses no such risks.

\item {\bf Licenses for existing assets}
    \item[] Question: Are the creators or original owners of assets (e.g., code, data, models), used in the paper, properly credited and are the license and terms of use explicitly mentioned and properly respected?
    \item[] Answer: \answerYes{} 
    \item[] Justification: The paper properly credits the datasets, models, and prior works used in the experiments, and uses resources from their official releases.

\item {\bf New assets}
    \item[] Question: Are new assets introduced in the paper well documented and is the documentation provided alongside the assets?
    \item[] Answer: \answerYes{} 
    \item[] Justification: The released code and accompanying documentation are provided in the supplementary material.

\item {\bf Crowdsourcing and research with human subjects}
    \item[] Question: For crowdsourcing experiments and research with human subjects, does the paper include the full text of instructions given to participants and screenshots, if applicable, as well as details about compensation (if any)? 
    \item[] Answer: \answerNA{} 
    \item[] Justification: We do not conduct crowdsourcing experiments or research involving human subjects.

\item {\bf Institutional review board (IRB) approvals or equivalent for research with human subjects}
    \item[] Question: Does the paper describe potential risks incurred by study participants, whether such risks were disclosed to the subjects, and whether Institutional Review Board (IRB) approvals (or an equivalent approval/review based on the requirements of your country or institution) were obtained?
    \item[] Answer: \answerNA{} 
    \item[] Justification: We do not conduct research involving human subjects or collect participant data.

\item {\bf Declaration of LLM usage}
    \item[] Question: Does the paper describe the usage of LLMs if it is an important, original, or non-standard component of the core methods in this research? Note that if the LLM is used only for writing, editing, or formatting purposes and does \emph{not} impact the core methodology, scientific rigor, or originality of the research, declaration is not required.
    \item[] Answer: \answerYes{} 
    \item[] Justification: We use multimodal foundation models as core components of the proposed framework and clearly describe their roles and usage in the paper.

\end{enumerate}

\end{document}